\documentclass[manuscript,screen]{acmart}

\usepackage{xcolor}
\usepackage[normalem]{ulem}

\usepackage[linesnumbered,ruled,vlined]{algorithm2e}
\makeatletter
\newcommand{\removelatexerror}{\let\@latex@error\@gobble}
\makeatother

\usepackage{url}
\usepackage{subcaption}
\usepackage{multirow}
\usepackage{tabularx}
\usepackage[bb=dsserif]{mathalpha}
\usepackage{cancel} 

\SetCommentSty{mycommfont}

\AtBeginDocument{%
  \providecommand\BibTeX{{%
    \normalfont B\kern-0.5em{\scshape i\kern-0.25em b}\kern-0.8em\TeX}}}

\setcopyright{acmcopyright}
\copyrightyear{2018}
\acmYear{2018}
\acmDOI{XXXXXXX.XXXXXXX}

\begin{document}

\title{Knowledge Equivalence in Digital Twins of Intelligent Systems}


\author{Nan Zhang}
\authornote{Corresponding Author.}
\email{zhangn2019@mail.sustech.edu.cn}
\affiliation{%
  \institution{Southern University of Science and Technology (SUSTech)}
  \city{Shenzhen}
  \country{China}
}
\affiliation{%
  \institution{University of Birmingham}
  \city{Birmingham}
  \country{UK}}

\author{Rami Bahsoon}
\affiliation{%
  \institution{University of Birmingham}
  \city{Birmingham}
  \country{UK}}
\email{r.bahsoon@cs.bham.ac.uk}

\author{Nikos Tziritas}
\affiliation{%
  \institution{University of Thessaly}
  \country{Greece}
}
\email{nitzirit@uth.gr}

\author{Georgios Theodoropoulos}
\authornotemark[1]
\affiliation{%
  \institution{Southern University of Science and Technology (SUSTech)}
  \city{Shenzhen}
  \country{China}
}
\affiliation{%
  \institution{Research Institute for Trustworthy Autonomous Systems}
  \city{Shenzhen}
  \country{China}
}
\email{theogeorgios@gmail.com}





\renewcommand{\shortauthors}{Zhang, et al.}

\begin{abstract}
A digital twin contains up-to-date data-driven models of the physical world being studied and can use simulation to optimise the physical world. However, the analysis made by the digital twin is valid and reliable only when the model is equivalent to the physical world. Maintaining such an equivalent model is challenging, especially when the physical systems being modelled are intelligent and autonomous.
The paper focuses in particular on digital twin models of intelligent systems where the systems are knowledge-aware but with limited capability. The digital twin improves the acting of the physical system at a meta-level by accumulating more knowledge in the simulated environment. 
The modelling of such an intelligent physical system requires replicating the knowledge-awareness capability in the virtual space.
Novel equivalence maintaining techniques are needed, especially in synchronising the knowledge between the model and the physical system. 
This paper proposes the notion of knowledge equivalence and an equivalence maintaining approach by knowledge comparison and updates.
A quantitative analysis of the proposed approach confirms that compared to state equivalence, knowledge equivalence maintenance can tolerate deviation thus reducing unnecessary updates and achieve more Pareto efficient solutions for the trade-off between update overhead and simulation reliability.
\end{abstract}



\maketitle

\section{Introduction}\label{sec:introduction}
Digital Twins (hereafter referred to as DT) is a concept that comes from manufacturing and is now gaining attention in other fields including logistics, transportation, city and urban planning, infrastructure monitoring and smart agriculture \cite{Jones2020CharacterisingReview,Minerva2020, DosSantos2021,Barricelli2019,Ghandar_2021}.
A Digital twin is generally regarded as the virtual representation of existing physical assets, products, objects, or systems.
A large number of definitions of DT have been proposed in the literature, each of which is strongly based on application context \cite{Barricelli2019,thelen_comprehensive_2022}. This paper focuses on DT that is used for runtime decision-making and planning \cite{DosSantos2021,flammini_digital_2021}. 
A representative and inclusive definition of DT, which is adopted in this paper, has been put forth in \cite{arup}: “A digital twin is the combination of a computational model and a real-world system, designed to monitor, control and optimise its functionality. Through data and feedback, both simulated and real, a digital twin can develop capacities for autonomy and to learn from and reason about its environment.”  Research into DT is still in its infancy and much of the work has been for prototyping and proof-of-concept purposes, covering but not limited to modelling, simulation, verification, validation, and/or assurance of the physical artefact using the virtual replica. The focus of the analysis has been 
geared towards simulation and what-if analysis to inform deployment and subsequent refinements of the physical products or systems through continuous monitoring and data assimilation \cite{dehghanimohammadabadi_simulation-optimization_2021, Tao_product_2019, vanderhorn_digital_2021, Boschert2016}.

One class of systems that can benefit from Digital Twin is intelligent systems, where the Digital Twin can assist in empowering intelligence in the physical system \cite{Zhang2020}. By intelligent systems, we refer to systems composed of intelligent agents \cite{wooldridge_jennings_1995}, whose intelligence is characterised by the ``self-*'' (self-awareness, self-configuring, self-healing, self-optimising, self-protecting, etc.) adaptive capabilities \cite{huebscher_survey_2008, Salehie2009}. 
An example would be that of a large-scale autonomous system consisting of connected computing nodes (e.g. drones), each of which is managed by an onboard intelligent agent. Each node can accumulate knowledge about itself and the environment to make its own decisions and to collaborate with other nodes in the network, where processing can be distributed and the control can be decentralised to each agent \cite{Diaconescu2017ArchitecturesSystems,Cardellini2018,Weyns2013}. The degree of autonomy and level of intelligence of these agents may be limited by issues like energy consumption and computation capability of each node. To overcome these limitations and enable better autonomy in the physical system, the agents of the physical system can be assisted by DT.
In \cite{Zhang2020} the authors have introduced the concept of self-aware DT and outlined an architecture for cognitive twins that can make informed decisions via predictive and prescriptive what-if analysis in the simulation environment leveraging knowledge awareness and cognition capabilities. The main question examined in \cite{Zhang2020} is whether and how the DT can be utilised to enhance the intelligence in the physical system. This paper aspires to address a different challenge, namely how best to decide the equivalence between a cognitive digital twin and its self-aware physical counterpart. To address this challenge, the architecture presented in \cite{Zhang2020} is revisited, generalised and enhanced with equivalence checking capabilities.


Predictions and analysis made by the digital twin can easily become obsolete when the physical world is highly dynamic and evolves in real-time, and when the DT knowledge drifts, decays and/or is not updated to reflect on changes in the physical world. The requirement for high fidelity of the digital twin system poses an important challenge to decide at which point a twin can indeed be considered a replica and not a mere abstract representation model of the physical system. In other words, fidelity replication begs the question: when is a model considered equivalent to the real system, and therefore the model can be trusted, and no further updates are required? Too frequent updates in the model introduce excessive communication and computation overheads, while large delays in updating the model force it to drift from reality \cite{DosSantos2021}. This paper aims to address this question in the context of DT of intelligent systems, from the perspective of knowledge.

A DT is essentially a data-driven info-symbiotic system, where data assimilated from the physical system(s) steer the computation and control of these systems to provide prescriptive, prognostic diagnostics and/or predictive analysis of the physical system \cite{Zhang2020}. The info-symbiotic relationship between the two worlds can generate a wealth of knowledge over time. Knowledge synchronisation and equivalence become of paramount importance and prerequisite for meaningful, effective, and efficient control and analysis, as knowledge drift and decay can make the twining obsolete. Additionally, maintaining knowledge equivalence can subsume other fundamental issues that can be rooted in data sources, sensing, processing, actuating, control, etc.  This can be particularly important in DT, supporting intelligent systems that are grounded on knowledge, covering dimensions that relate to stimuli, time, interaction, goals, etc. 

Current approaches for building Digital Twin do not provide explicit solutions to the problem of knowledge equivalence as a fundamental problem for ensuring fidelity between the physical and digital worlds. 
This paper specifically looks at knowledge equivalence in the design of intelligent systems, assisted by the DT. In particular, it investigates the challenge of real-time modelling of autonomous systems which can accumulate knowledge through DT to support intelligence and how knowledge equivalence can be engineered and/or reasoned about in this setting. 
 
The work presented in the paper assumes that a digital twin is an imperfect artefact  when attempting to ``follow'' the physical system, and  calls for knowledge comparison between the digital and physical worlds \cite{Gao2021AnSystems}.
The agents operating in the physical world are considered \textbf{interaction-aware}, that is, being able to learn models capturing knowledge specifically about their interaction with other agents and the environment. The knowledge of interaction is modelled as a weighted graph by each agent.
The objective is (i) to inform when and how often to update the DT knowledge, and (ii) what is an acceptable tolerance and knowledge deviation level between the digital and physical worlds. The paper posits that utilising knowledge equivalence as opposed to \textit{state equivalence} can alleviate the overheads while improving simulation validity and can be a more efficient and effective approach for DT of intelligent systems.
The contribution of this paper is as follows:
\begin{itemize}
    \item A novel reference model for the design of digital twins (DT) of intelligent systems, where knowledge  equivalence is a key component.
    \item The formulation of the problem of knowledge equivalence for the DT of an intelligent physical system.  This includes an outline of the threats to knowledge equivalence and their detection and impact as well as a set of metrics that measure the equivalence of fine-grained knowledge related to interaction. 
    \item An online knowledge equivalence checking framework that identifies discrepancies  and determines  when to update the DT.
    \item Demonstration and quantitative evaluation of the proposed framework based on a smart camera network scenario where agents are interaction-aware.
\end{itemize}



The rest of the paper is organised as follows. 
Section \ref{sec: background} first presents the background and related work.
Section \ref{sec: DT modelling} provides a reference architecture for  DT of intelligent systems that can incorporate different levels of knowledge and self-awareness.
Section \ref{sec:equivalence} provides a theoretical formulation of the problem of knowledge equivalence, and outlines relevant threats and metrics. The proposed methodology for online knowledge equivalence checking is presented in Section \ref{sec:checking}. Section \ref{sec:evaluation} presents a prototype implementation and experimental evaluation of the proposed approach.  Section \ref{sec:conclusion} concludes the paper and outlines future research directions.

\section{Background And Related Work} \label{sec: background}

\subsection{Digital Twins} \label{subsec: DT}

The specific definition of DT varies in different application domains, but in general referring to the models that link with physical assets, objects, or products in real-time or throughout the full life-cycle of the physical entities. 
One of the earliest and most fundamental documented definitions of DT is given by NASA in 2012 as an initiative for future generations of vehicles operating in extreme service conditions:
``A Digital Twin is an integrated multiphysics, multiscale, probabilistic simulation of an as-built vehicle or system that uses the best available physical models, sensor updates, fleet history, etc., to mirror the life of its corresponding flying twin \cite{glaessgen_digital_2012}.''
With the simulation predictive capability, the DT can forecast the system health, remaining useful life, and system response in previously unknown conditions.
Tao et al. proposed a five-dimensional model of DT in the manufacturing domain, which defines five components of the DT: physical part, virtual part, connection, data, and service \cite{tao_digital_2017, tao_digital_2019_8049520}. 
In \cite{mihai_digital_2022}, the authors further add to the definition of DT with the capabilities of self-adapting, self-regulating, self-monitoring, and self-diagnosing.
A taxonomy of DT is also presented in \cite{van_der_valk_taxonomy_2020}.

The emphasis of DT research has been recently put on the dynamic and bi-directional connection between the DT and the system being modelled, and how to link the physical object and digital object in an accurate and real-time manner \cite{liu_review_2021}. In the Internet of Things (IoT) context, \citet{Minerva2020} proposes a set of essential properties that characterise the DT, among which ``the constant entanglement between an artifact and its software representations'' is of prominent importance to reflect changes of the physical artifact to the model in real-time and vice versa. 
The directions of data flow between the DT and the physical object/entity are classified as virtual-to-physical (V2P) and physical-to-virtual (P2V) \cite{thelen_comprehensive_2022}. In P2V, the state of the model is updated with sensor data. In V2P, the up-to-date model is used to predict the future state of the physical system and actuate decisions back to the physical system.

A concept similar to DT is Dynamic Data-Driven Applications Systems (DDDAS), which is first introduced by Frederica Darema. DDDAS describes the ability to ``incorporate additional data into an executing application (these data can be archival or collected online), and in reverse, the ability of applications to dynamically steer the measurement process'' \cite{darema_dynamic_2004}.
The core of DDDAS is the data assimilation and the sensor reconfiguration computational feedback loops. The data assimilation loop dynamically incorporates extra data into the model, where errors between the simulation prediction and the sensor data from the physical system are used to refine the model, enabling the simulation to follow the trajectory of the physical system. In the sensor reconfiguration loop, the application simulation will, in return, adjust the sensing and measurement (e.g. sampling a subset of the entire measurement space) for better modelling and prediction purposes  \cite{blasch_dddas_2018}. 
The DDDAS paradigm can be an enabling foundation for DT in the design of feedback loops between the physical system and its virtual model \cite{diamantopoulos_digital_2022, Zhang2020, Ghandar_2021, kapteyn_probabilistic_2021,malik_dynamic_2020}.

To support accurate simulations by DT, models that are equivalent to the physical world are needed. This paper focuses on the DT modelling of intelligent systems. 
The next subsection delineates the scope of intelligent systems in this paper.

\subsection{Intelligent Self-Aware Systems}\label{subsec: intelligent systems}

According to IEEE standard 7010-2020,
An \textit{autonomous/intelligent system (A/IS)} is defined as ``a semi-autonomous or autonomous computer-controlled system programmed to carry out some task with or without limited human intervention capable of decision making by independent inference and successfully adapting to its context. An example is an A/IS that refers to a computer system instantiated in a product or service'' \cite{ieee_std_7010-2020}.
This definition correlates autonomy to the intelligence of systems. Further, an \textit{autonomous system} defined by IEEE standard 7001-2021 refers to ``a system that has the capacity to make decisions itself in response to some input data or stimulus with a varying degree of human oversight or intervention depending on the system’s level of autonomy'' \cite{ieee_std_7001-2021}.
Therefore, the emphasis of intelligence is the system's capability to independently plan for its behaviour in a dynamic environment.

In designing intelligent capabilities, IBM proposed the concept of \textit{autonomic computing} in 2001, which compares complex computing systems to the human autonomic nervous systems and promotes self-governing  properties to be incorporated into the system design \cite{huebscher_survey_2008,ibm_autonomic_2001}. 
The vision was put forth under the increase in the number of interconnected computing devices that operate cooperatively. It is infeasible for system administrators to anticipate the behaviour of interactions, thus requiring runtime adaptive capabilities for the configuration, optimisation and maintenance of such complex systems \cite{kephart_vision_2003}. 
An autonomic computing system should have self-management capabilities to change in accordance with business policies and objectives, such as: self-configuration, self-optimisation, self-healing, and self-protection \cite{ibm_architectural_2005}. 
Each system component is architected with a control loop denoted as MAPE-K (Monitor-Analyse-Plan-Execute with Knowledge) \cite{ibm_architectural_2005}.

The manageability problem under complexity that autonomic computing aims to tackle motivates the study of self-adaptive systems \cite{weyns_introduction_2021}.
\textit{Self-adaptation} as an intelligent ``self-*'' capability is also highly influenced by the MAPE-K architecture.
A self-adaptive system uses closed-loop control to adjust its behaviour in response to the dynamic environment and the awareness of itself \cite{cheng_software_2009}.

The knowledge repository in the MAPE-K loop and self-adaptation, in general, is an important component for decision-making and is central to intelligent systems. The maintenance of such a knowledge repository is specifically addressed by \textit{self-aware computing} \cite{lewis_self-aware_2016}. \textbf{\textit{Self-awareness}} can be regarded as the lowest level and an enabler of self-adaptation and other self-* properties, meaning the system is aware of the state of itself and environment contexts, in order to extract, accumulate and maintain knowledge through interaction with the environment and other computing devices \cite{Salehie2009}. With the knowledge repository, a self-expression process can deliberately plan and decide the next actions to self-adaptively change the system's behaviour \cite{lewis_self-aware_2016}.

Knowledge involves different cognitive dimensions and can be captured in different representations. 
The paradigm of computational self-awareness draws inspiration from human cognition and levels of awareness in the human brain to represent knowledge of each system component  \cite{Lewis2015,chen2015handbook,Zhang2020}. 
A self-aware system can pose four dimensions of knowledge thus possessing capabilities manifested as four levels of self-awareness: stimulus-, interaction-, time-, and goal-awareness to reason about and formulate its actions:
\begin{itemize}
    \item \textit{Stimulus-awareness}: Stimulus-awareness is the basis for all levels of self-awareness and describes the simple reactive behaviour of the system. Stimulus-awareness is the ability to perceive the stimuli or events acting on the system and can consequently evoke the system to decide its next action. 
    \item \textit{Interaction-awareness}: An interaction-aware system can perceive historical knowledge of state, behaviour, and type of relationships and communication of interacting system components; it can consequently use this knowledge to decide its subsequent actions. Interaction-related knowledge may be represented as a graph or network among other representations. 
    \item \textit{Time-awareness}: Being aware of time means the system is able to use models that involve explicit memory, time-series modelling, or anticipation \cite{Lewis2015}.
    \item \textit{Goal-awareness}: Goal-awareness refers to the ability to know the system goals and reason about the goals. The awareness can be expressed as the utility or satisfaction of goals.
\end{itemize}

Different levels of self-awareness can exist within the system at the same time, but stimulus-awareness (i.e, as the basic awareness level) is the prerequisite and enabler for other awareness levels.

This paper specifically refers to an intelligent system as a system that is composed of interconnected computing devices. Each computing device is controlled by an onboard autonomous agent with self-awareness capability. Each self-aware agent is assumed to be \textbf{interaction-aware}, accumulating knowledge by modelling its interaction with the environment and other agents. Each agent maintains its own knowledge base and uses the interaction model to plan its actions.

\subsection{Related Work}
\label{sec:related work}
\subsubsection{DT for Intelligent Systems}
Designing DT of  systems that are already embedded with intelligent self-* capabilities is still at a preliminary stage. 
Much of the work involving architecture/reference model design focuses on offloading the intelligence that originally can be run on the physical system to the DT. 
In \cite{Zhang2020}, we have proposed a centralised cognitive DT reference model that incorporates self-awareness capabilities that accumulate, revise, and utilise knowledge about the real world and the DT itself at different fine-grain levels (stimulus, interaction, time, goal) to inform the optimisation of the behaviour of multi-agent-based systems. 
Effort has been made in architecting DT for autonomous robots with co-simulation \cite{lumer-klabbers_towards_2021}. The work in \cite{lumer-klabbers_towards_2021} proposes an architecture to delineate the components specific to the control of robot applications.  The decision-making capability of the robot is offloaded to the DT which supports movement prediction and emergency stops.
There is also research work that studies collective DTs for systems that are composed of collective nodes.
The work in \cite{Rivera_forging_2022,Rivera2020OnSystems} proposes a framework for autonomic and cooperating DTs with the integration of dynamic context management, runtime models, adaptive control, and feedback loops. The framework contains dynamic context management by the MAPE-K (Monitor-Analyze-Plan-Execute-Knowledge) loop which is widely adopted by self-adaptive systems. Each DT is able to run experiments to update its knowledge to achieve its goals, while all DTs can self-organise for a cooperative reaction to a shared goal.
Casadei et al. coined the notion of ``augmented collective digital twin'' for self-organising cyber-physical systems, which fuses the concept of DT, virtual devices, and collective systems \cite{casadei_augmented_2021}. They have designed a meta-model that maps logical devices and their internal components to physical nodes, i.e. where the components of the logical device are deployed. The self-organising behaviour happens among the logical devices in the virtual space.
in \cite{ricci_web_2022}, the notion of Web of Digital Twins (WoDT) is proposed to envision a distributed ecosystem of connected DTs. The physical system is assumed to be composed of physical assets in a broad meaning that can refer to physical objects, persons, or processes. A state manager component is involved in the architecture to update the state of DT with events dynamically captured from the physical assets. 

To summarise, most existing efforts in augmenting the self-* intelligent capabilities of the physical system have been put forth in offloading full intelligence to the DT.  An assumption not adequately explored is that a limited level (due to resource constraints) of intelligence and autonomy residing in the physical twin and the DT  augments that intelligence with what-if simulation analysis. The basis of accurate simulation is an equivalent model in the DT. Further, when modelling physical systems that are able to self-learn, additional requirements should be raised to ensure the knowledge of the physical system is also equivalently maintained in the DT. One of the contributions of this paper is an architecture that addresses the above-mentioned problem of knowledge equivalence, which is not discussed in the existing literature.

\subsubsection{Equivalence in DT}

The concept of equivalence has been investigated in different domains and problems including 
model verification and validation \cite{Sargent2010VerificationModels,Carson_model_validation2022,DoD}, 
dynamical systems \cite{Schaft2004}, 
model checking \cite{Biere1999,Clarke2012}, 
program equivalence \cite{Sharma2013, Churchill2019},
software evolution \cite{Wang2018VerifyingApplications},
knowledge representation \cite{Giles1999},
bisimulation of concurrent systems \cite{DeFrancesco2016},
agent-based model adaptation \cite{Gronauer_2022},
VLSI design\cite{Bombieri2007,Zhu_logic_equivalence_2008}, and
industrial control systems \cite{wang_equivalence_2008}.
Grounding models to data in an online feedback loop has been extensively investigated in the context of DDDAS systems (as mentioned earlier in Section \ref{subsec: DT}), a comprehensive review of approaches to implement the feedback loop is provided in \cite{blasch_dddas_2018}. 

The rest of the subsection provides an overview of approaches and techniques that focus specifically on the issue of twinning addressing the problems of how to decide when the model needs to be updated and how the model is updated. Table \ref{table_review} summarises these approaches. 
Three main classes of approaches are distinguished, namely replicating the sensor data without any comparison; explicitly identifying discrepancies between the model and the real system; continuous model calibration. 
With regard to model adaptation, four elements that may be updated are state variables, parameters, behavioural rules and input. 








\begin{table}[!t]

\caption{Related work for online equivalence}
\label{table_review}
\centering
\scalebox{0.85}{
\begin{tabularx}{\textwidth}{ccccX}
\toprule
Category & Reference & Comparison entities & Update & Method \\
\midrule
Replication & \cite{Ambra2020Agent-BasedSpaces,Korth2018,Clemen2021Multi-AgentCities} & - & State & \\
Replication & \cite{Eckhart2018ATwins} & - & Input (stimuli) & Stimuli identification \\
Equivalence Checking & \cite{kennedy_aimss_2007} & Knowledge (rules of agents) & - & Association rule mining \\
Equivalence Checking & \cite{Gao2021AnSystems} & Outputs given the same inputs & - & Gaussian Mixture Model \& Hidden Markov Model \\
Equivalence Checking & \cite{hashemi_multi-agent_2017} & State & Parameter & Reinforcement learning \\
Equivalence Checking & \cite{lugaresi_online_2022} & Output sequence &  Parameter & Online validation by time series analysis \\

Online Calibration & \cite{papathanasopoulou_online_2016,hammit_case_2018} & State & Parameter & Optimisation \\
Online Calibration & \cite{Zipper2019SynchronizationTwin} & Output & Input & Optimisation \\
Online Calibration & \cite{Henclewood2012} & Statistical test on state & Parameter & Optimisation \\
Online Calibration & \cite{kapteyn_probabilistic_2021} & N/A & State & Bayesian inference \\
Online Calibration & \cite{akbarian2020synchronization} & N/A & N/A & Kalman filter or PID \\
Online Calibration & \cite{Jia2020} & Output & Parameter & Error feedback \\
Online Calibration & \cite{Naing2021} & State & Parameter & Online training \\
\bottomrule
\end{tabularx}
}
\end{table}

\paragraph{Sensor Data Replication}
This approach aims at achieving equivalence by directly replicating sensor data into the model. 
The sensor data may be used to update the state of the model either at a fixed frequency \cite{Ambra2020Agent-BasedSpaces} or dependent on the availability of data or event \cite{Korth2018,Clemen2021Multi-AgentCities}. The aim is to ensure the state of the model reflects that of  the real system. 
The work in \cite{tan_optimizing_2022} studied adaptive synchronisation for balancing the trade-off between simulation accuracy and the costs associated with frequent updates.
The problem is formulated as finding an optimal control policy that decides when to synchronise the state of the DT. However, their work assumes DT has an accurate model of the physical system, which hinders the applicability to online decision-making under a dynamic environment.

In industrial control systems, a state usually refers to a hidden property of the system that cannot be obtained through sensors. Sensors can only measure the input and the resulting response (output) of the system in order to estimate the current state.
In  \cite{Eckhart2018ATwins} the state is replicated by identifying and updating the  stimuli inputs in industrial control systems. 
This work assumes the physical system and the twin system run the same program defined by a finite state machine. However, the state transition of the program is fixed and does not evolve over time, indicating no learning and awareness of knowledge are involved.

Our work instead focuses on twinning a system that maintains and evolves its own knowledge base. State replication alone is not enough to allow the DT to ``mirror'' all the perspectives and dynamics of the system. Real-time replication of the knowledge base is also needed for intelligent systems.

\paragraph{Discrepancy Checking}

This approach focuses on explicitly comparing the model and the real system to identify discrepancies and decide whether and when to update the model.
The AIMSS system has investigated this problem in the context  of info-symbiotic (DDDAS) agent-based social simulations\cite{doi:10.1260/1748-3018.5.4.561,10.1007/11758532_74,kennedy_aimss_2007}. The system utilises a SAT solver to  detect inconsistencies of different rule sets from the model and the real world dynamically identified via associative rule mining. The rules can be viewed as knowledge of the agents, however, this is performed offline.
The discrepancy detector in  \cite{Gao2021AnSystems} utilises a Gaussian mixture model to compare the outputs of the real system and model given the same inputs and detect anomalies due to the missing mode of the model.
Checking the inconsistency of states using reinforcement learning (RL) agents is described in  \cite{hashemi_multi-agent_2017}. The RL agents take the state difference as inputs and decide whether to update and which parameter to calibrate.
In \cite{lugaresi_online_2022}, the authors propose an online validation method for DT of manufacturing systems. Dynamic Time Warping, which is a time series analysis algorithm, is used to compare the inter-departure time sequences of the manufacturing queue produced by the physical system and a Discrete Event Simulation model. A threshold is used to alert the validity of the model. If not valid, the parameters for the distribution of processing time are re-fitted with the recently acquired data.

Existing effort in checking the equivalence between the DT and the real system is limited to the comparison of states, input or output. There is still a lack of attention to knowledge equivalence checking and comparison in order to identify whether the DT maintains a virtual intelligent agent that possesses knowledge equivalent to the knowledge of the real-world intelligent agent. As explained in section \ref{sec:introduction} and \ref{subsec: intelligent systems}, this is particularly important for intelligent systems.
\paragraph{Continuous Online Calibration}
Continuous online model calibration is also utilised for maintaining an equivalent model. Calibration focuses on continuously tuning the model parameters, typically at a fixed frequency.
Approaches in this category include the use of:  difference/error between observed and predicted values of the state variables as the objective function to optimise the selection of parameters or inputs  \cite{papathanasopoulou_online_2016,hammit_case_2018,Zipper2019SynchronizationTwin}; statistical tests on simulated and real traffic flows as calibration criteria \cite{Henclewood2012}; Bayesian inference in the context of unmanned aerial vehicle applications \cite{kapteyn_probabilistic_2021}; Kalman filter  \cite{akbarian2020synchronization}; the error between observed output and predicted output as feedback for the calibration of clock models \cite{Jia2020};  and online calibration by training deep learning models with streaming data (state variables) \cite{Naing2021}.


\section{Digitally Twinning Intelligent Systems}
\label{sec: DT modelling}

Twining intelligent systems poses challenges and opportunities. In particular, implementing intelligence in the physical world can be constrained by limitations in the physical environment, such as space, computation, power, memory, mobility, coordination and/or unavailability of knowledge. Henceforth, the self-* intelligent capability can be limited. Twining can be architected to provide support scenarios, for ``empowering intelligence'' to overcome these limitations. Consequently, knowledge management and equivalence in DT can become vital, to unlocking new modes of intelligence. 

This section defines the real-world intelligent system considered in this paper, and proposes a solution on how the DT models such a system as well as the environment in which the system operates. A novel reference architectural model for DT is presented later to address the knowledge equivalence management and optimisation empowered by DT.
For the clarity of terminology, throughout the rest of the paper, the system that operates in the physical environment is called the \textit{physical system}. A physical system is composed of agents, which are referred to as \textit{physical agents}. Correspondingly, the DT model of a physical system is called a \textit{virtual system}. The model of a physical agent is a \textit{virtual agent}. The model of a physical environment is called a \textit{virtual environment}.

\subsection{Physical World Model}

\begin{figure}
\centering
\begin{minipage}[b]{.45\textwidth}
  \centering
  \includegraphics[width=\linewidth]{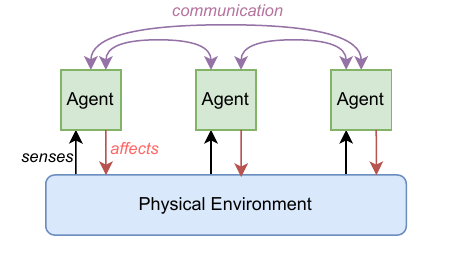}
  \captionof{figure}{Physical world.}
  \label{fig: physical_world}
\end{minipage}%
\begin{minipage}[b]{.53\textwidth}
  \centering
  \includegraphics[width=\linewidth]{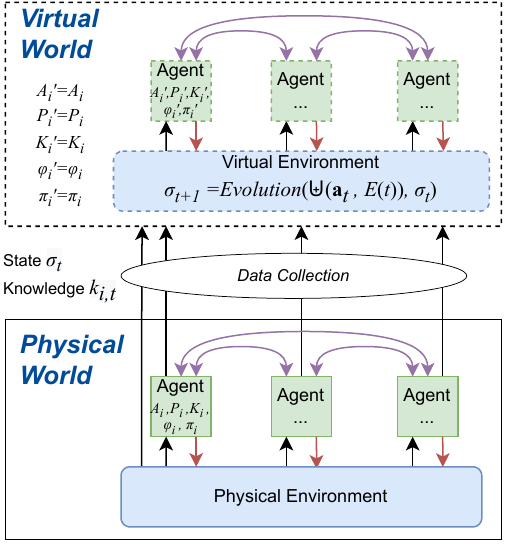}
  \captionof{figure}{Modelling of the physical world.}
  \label{fig: physical-virtual}
\end{minipage}
\end{figure}

Consider a scenario where the physical system is composed of computing nodes with embedded intelligence. Each node is controlled by an onboard intelligent self-aware agent (referred to as a \textit{ physical agent}) with its own local knowledge about itself and the environment. Agents collaborate and accumulate knowledge as shown in Fig. \ref{fig: physical_world}.    
Such a configuration is common in the studies of decentralised self-adaptive/self-aware systems \cite{Weyns2013,quin_decentralized_2021}. 
For instance, in surveillance applications, geologically distributed smart cameras can be deployed to track objects of interest as they move through the environment. The cameras self-organise the tracking handover without central control or prior knowledge of the camera network topology. Each camera as a computing node is embedded with an agent that communicates with other cameras to negotiate the tracking responsibility. 
Each agent can be designed with time- and interaction-awareness: adaptively learning knowledge of interaction from successful handovers with other cameras in the past to infer the camera network topology, which can be used for handover negotiations \cite{Esterle2014Socio-economicNetworks}. 
Another example is decentralised elastic data stream processing in the cloud \cite{CARDELLINI2018171}, where operator programs running on server nodes need to process incoming data streams in multiple steps to get the output. To speed up the processing while minimising reconfiguration costs, each operator instance can be scaled in/out and migrated to other nodes.  Each operator can be managed by a self-aware agent individually. Each agent adaptively learns the resource utilisation pattern and decides the scaling and migration of each operator autonomously, which is more scalable compared to a centralised control scheme.

\subsubsection{Intelligent Self-Aware Agents} 
\label{sec:agent}
The physical agents in this paper are considered to possess self-awareness capability.
An agent program is a mapping from perception to action \cite{AIbook}. Inspired by the concept of \textit{learning agents} \cite{AIbook} and \textit{self-awareness} \cite{Lewis2015}, we further formally specify the intermediate process between perception and action by explicitly incorporating the awareness and utilisation of knowledge.

An intelligent physical agent $\alpha$ is regarded as a tuple $\{ A, P, K, \phi, \pi\}$, where:
\begin{itemize}
    \item $A$ is a set of all allowed actions of an agent. An agent affects the environment via actions. Communication between agents is also viewed as actions \cite{wooldridge_jennings_1995}. 
    \item $P$ is the agent's perception space of its environment and itself. The perception at time $t$ is denoted as $p_t \in P$, which consists of processed data from the sensors. Example content of the perception data may include the received messages from other agents or the positions of the agent's surrounding objects in the environment.
    \item $K$ is the space of the agent's runtime knowledge, with $k_t\in K$ signifying the agent's knowledge base at time $t$; Knowledge is represented according to the \textit{computational self-awareness} architecture, explained in section \ref{subsec: intelligent systems}. 
    \item $\phi$ is the agent's awareness function, which constructs the agent's runtime knowledge $k_t$  based on the perception  $p_t$  and its previous knowledge $k_{t-1}$  (supposing time is discrete). $\phi$ can be used to instantiate the learning process in which $p_t$  is the feedback after utilising $k_{t-1}$ .
    \begin{equation}
        k_t = \phi(p_t, k_{t-1}) \label{eqn: awareness}
    \end{equation}  
    It is worth noting that if the agent is only stimulus-aware, its awareness function $\phi$ will degenerate from $\phi: P \times K \rightarrow K$ as specified by equation (\ref{eqn: awareness}) to $\phi: P \rightarrow K$, because stimulus-awareness alone is purely reactive and cannot preserve history. 
    \item $\pi$ is the agent's decision function. An agent utilises $\pi$ to decide the next action $a_t$  based on its perception $p_t$  and current knowledge $k_t$.
    \begin{equation}
        \pi: P \times K \rightarrow A,\ a_t = \pi(p_t, k_t) \label{eqn: decision}
    \end{equation}
    Notice that the output $a_t$ can be a single action or a probability distribution of different actions, which corresponds to deterministic and probabilistic decision-making, respectively.
    
\end{itemize}

The agent follows a sense-think-act cycle. During \textit{sensing}, the agent first obtains its perception $p_t$ by sensing its environment and itself, including checking the messages it receives from other agents (e.g. through a mailbox \cite{Chen2020}). In the \textit{think} process, the agent uses the current $p_t$ as feedback to update its knowledge by $\phi$. Then $\pi$ will derive an action or a probability distribution of actions. Finally, at the \textit{act} stage, the action $a_t$ is executed through actuators to affect the environment or/and other agents. 

Knowledge is different from an internal ``state''. 
A state within an agent is composed of a set of attributes that represents the current status of the agent or the current representation of the world. Examples of the internal state of an agent can be the energy consumption level, the current action, the agent's location in the environment, etc.  
However, knowledge refers to the set of beliefs and information that an agent has accumulated over time and that the agent uses to make decisions or take actions. It is often represented by a set of rules, models, or beliefs that describe relationships between variables and that are used by the agent to reason about its environment. These rules, models, or beliefs can be revised as the agent continuously interacts with the environment, which refers to the notion of online learning. Examples of knowledge may include the model of the environment dynamics, goals and how to achieve goals, perceived interaction patterns with other agents, etc. Knowledge supports the mapping from state to action.

Fig. \ref{fig: SA-agent} illustrates a self-aware agent based on the formal definition above and the self-awareness paradigm outlined in section \ref{subsec: intelligent systems}.
The self-awareness component constructs and modifies the knowledge base by its capabilities at one or multiple level(s) with sensor perceptions.
The \textit{self-expression} component expresses decisions and yields consequent actions entailed by the awareness level(s). 

\subsection{Virtual World}
The modelling of the physical world involves replicating agents and the physical environment as \textit{virtual agents} and \textit{virtual environment}, as shown in Fig. \ref{fig: physical-virtual}.
The behaviour of agents and the dynamics of the environment are modelled as follows.

\subsubsection{The Agent Model}
Since an agent is essentially a software system, the modelling towards a virtual agent can be realised by replicating the same software program of the physical agent. 
Specifically, a virtual agent possesses the same $\{A, P, K, \phi, \pi \}$ as its physical agent counterpart.

\subsubsection{The Environment Model}
To model the environment, we take a state-based perspective \cite{10.1007/978-3-540-32259-7_8,10.1145/2517449}. The environment is considered as a two-tuple: $Environment = <State, Process>$.
The environment has its own endogenous dynamics denoted by $Process$ that can change its $State$, independent of the actions of agents.
Let $\sigma \in \Sigma$ be the state of the environment. 
At any time $t$, the state of the environment $\sigma_t$ can be characterised by a vector that contains multiple variables:
\begin{equation}
\label{eqn: state}
    \sigma_t := ( \sigma_{1,t}, \sigma_{2,t}, ... \sigma_{k,t})
\end{equation}
For instance, in a spatial context, $\sigma_t$ can be composed of the positions of all physical entities physical agents, obstacles, etc. within the space. Variable $\sigma_{i,t}$  represents the position of the $i$-th entity.

The environment not only is affected by the agents but also evolves according to its own endogenous dynamics. 
The coupling of the environment and agents results in the overall world dynamics, which is defined by a function \textit{Evolution} such that the transition from current moment $t$ to the next time step $t+1$ is the composite result of all the agent actions and the endogenous dynamics of the environment \cite{Michel2009}:
\begin{equation}
    \sigma_{t+1} = Evolution(\uplus(\mathbf{a}_t, E_n(t)), \sigma_t) 
\end{equation}
$\mathbf{a}_t = (a_{1,t}, a_{2,t},...,a_{m,t})$  is all the action made by each agent onto the environment at time $t$, and $a_{i,t}$ is the action made by agent $i$ at time $t$.
$E_n(t)$ is the dynamics produced by the natural evolution of the environment (the $Process$).
The symbol $\uplus$ denotes the action composition operator, which defines how the actions and endogenous dynamics of the environment at $t$ are composed to calculate the consequences on the previous environment state $\sigma_t$ \cite{Michel2009}.

\subsection{A Reference Architecture}
\begin{figure}
\centering
\begin{minipage}[b]{.51\textwidth}
  \raggedright
  \includegraphics[width=\linewidth]{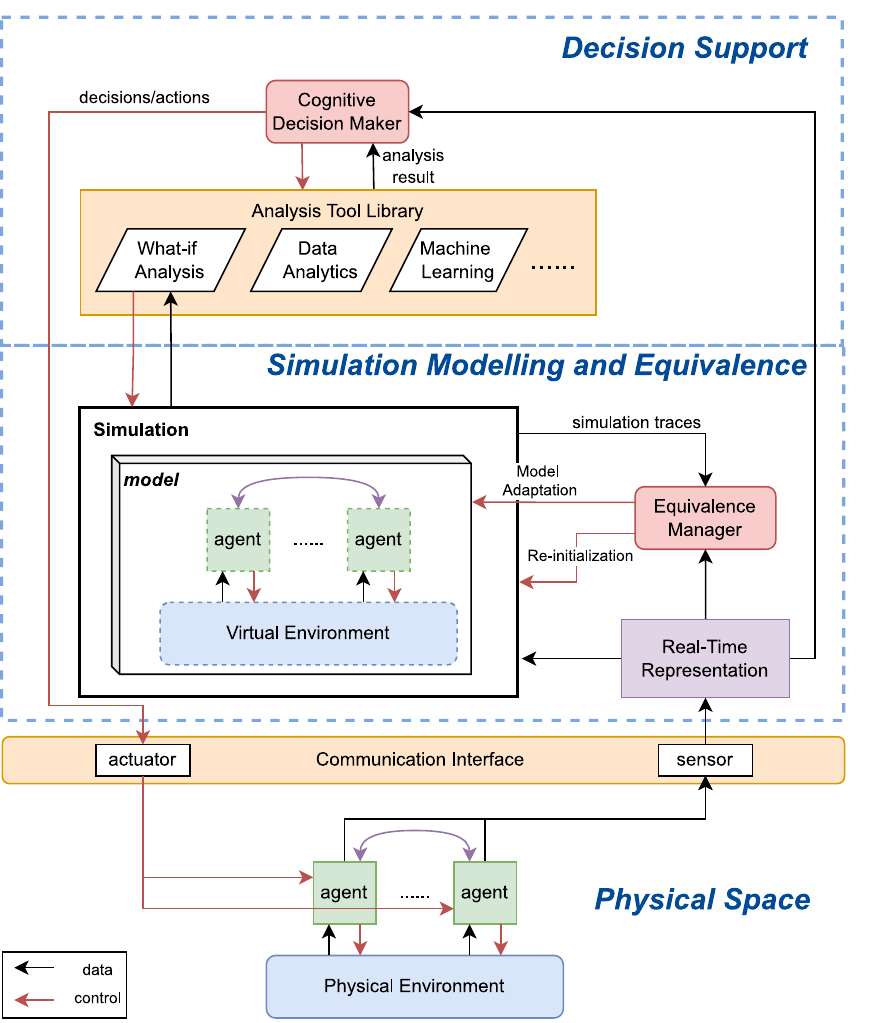}
  \captionof{figure}{A reference model of intelligent systems DT.}
  \label{fig: arch}
\end{minipage}%
\begin{minipage}[b]{.48\textwidth}
  \raggedleft
  \includegraphics[width=0.92\linewidth]{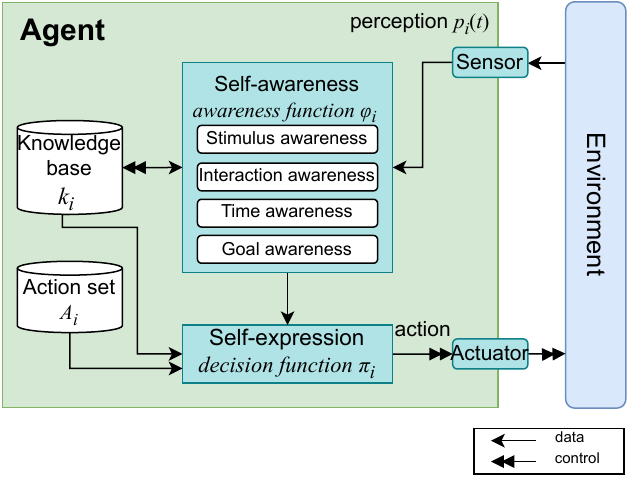}
  \captionof{figure}{A self-aware agent.}
  \label{fig: SA-agent}
\end{minipage}
\end{figure}

Different reference models of DT that enrich the intelligence of physical systems have been proposed as discussed in section \ref{sec:related work}.
However, these reference models do not involve intelligence in the physical system that is composed of multiple self-learning agents who evolve their own knowledge bases. The necessity of knowledge equivalence is not reflected in their designs, either. Based on the model abstractions in the previous section where the physical system is an intelligent multi-agent system, a reference model of a DT is proposed in Fig. \ref{fig: arch}.  
The DT and physical world interact through a communication layer. The data (perception and knowledge) sent from the agents in the physical world are first used to create the real-time virtual replica. Then these data are used to drive the model equivalence and decision support. 

\subsubsection{Cognitive Decision Support}

In the reference model, a Cognitive Decision Maker is introduced, which can be enriched by human cognitive behaviour of self-awareness to support predictive and prescriptive analysis \cite{Zhang2020}.
The cognitive decision maker monitors any event that is causing or may lead to sub-optimality of the physical world and then provide more informed decisions to optimise the behaviour of the physical system.
The analysis towards decisions is enabled by a library of analysis tools, among which the what-if analysis is of paramount importance.
With what-if analysis, the cognitive decision maker can obtain an in-depth understanding and explanation of the mechanisms governing the system's behaviour through simulation. By exploring different scenarios or experimenting with alternative decisions, the cognitive decision maker can deliver actionable insights suggesting the best solution given a variety of choices \cite{Zhang2020}.

The decision-making for the intelligent multi-agent system is performed at a meta-level. Since agents already have a certain level of intelligence and autonomy as described in Section \ref{sec:agent}, the DT mainly \textit{assists} the intelligence of the agents, for instance, on the agent's decision function $\pi$. An agent relies on its $\pi \in \Pi$ to derive an action, where $\Pi$ contains all possible decision functions that this agent can be equipped with. 
With a DT, the outcome of different candidates of $\pi$ can be evaluated in the simulation before enacting any changes in the physical world. Furthermore, by switching the level of knowledge, the DT can serve as an instantiation for the notion of meta-self-awareness \cite{Elhabbash2021}.
\subsubsection{Model Equivalence and Adaptation}
In the proposed framework, agents are assumed to be the only entities in the physical world that are able to transmit data to the virtual world.
Each agent $\alpha_i$ pushes its perception $p_{i,t}$  (or part thereof) and current knowledge $k_{i,t}$ to the virtual world, such that each virtual agent is updated with the most recent perception and knowledge, and the current environment state $\sigma_t$  can be interpreted by combining the perception from each agent.
The equivalence manager continuously compares the model against the physical world and, in case of discrepancies, it initiates additional data absorption or adaptation of the model \cite{doi:10.1260/1748-3018.5.4.561}. 

Traditionally, equivalence checking is largely based on state synchronisation and comparison between sensory data of the physical system and the output of the model or through the exhaustive exploration of transitions of states (see section \ref{sec:related work}). Such \textit{state equivalence} checking methods can suffer from heavy computational overheads, making them difficult to scale to large-scale settings, such as DT where equivalence should be maintained in real-time or near real-time.  This is exacerbated in highly dynamic applications where model adaptation is expected to be performed frequently to capture the up-to-date dynamics of the fast-evolving environment.

In  systems encompassing  intelligent agents, the dynamic non-deterministic behaviour of agents may also change the dynamics of the physical world and the world model. In such cases, the knowledge of the agents may become the determining factor affecting the deviation of these two respective components of the twin system as it is the knowledge of the agents that determine their decisions, their actions and their effect on their environment. Using knowledge as a metric for equivalence checking such intelligent systems emerges therefore as a viable approach that could overcome the high overheads involved in using  state, since knowledge is a task-related abstraction of state.

\subsubsection{Levels of Knowledge}
\label{Levels of Knowledge}
The agents in the reference architectural models are instantiated with the self-awareness paradigm. As introduced in section \ref{subsec: intelligent systems} and shown in Fig. \ref{fig: SA-agent}, a self-aware agent can maintain knowledge at different levels.
The self-awareness component constructs and modifies the knowledge base by its capabilities at one or multiple level(s) with sensor perceptions.
The \textit{self-expression} component expresses decisions and yields consequent actions entailed by the awareness level(s).

\subsubsection{Equivalence Manager }

\begin{figure}
\centering
\begin{minipage}[b]{.52\textwidth}
  \includegraphics[width=\linewidth]{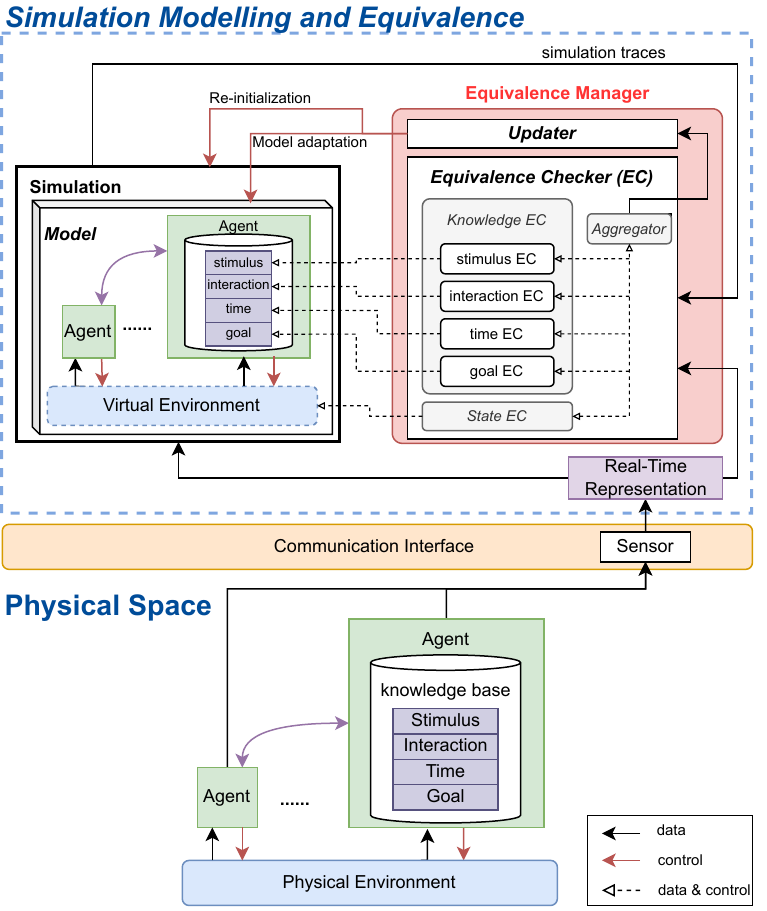}
  \captionof{figure}{Instantiation of Fig. \ref{fig: arch} with Self-Awareness.}
  \label{fig: arch-SA}
\end{minipage}
\hfill
\begin{minipage}[b]{.45\textwidth}
  \includegraphics[width=\linewidth]{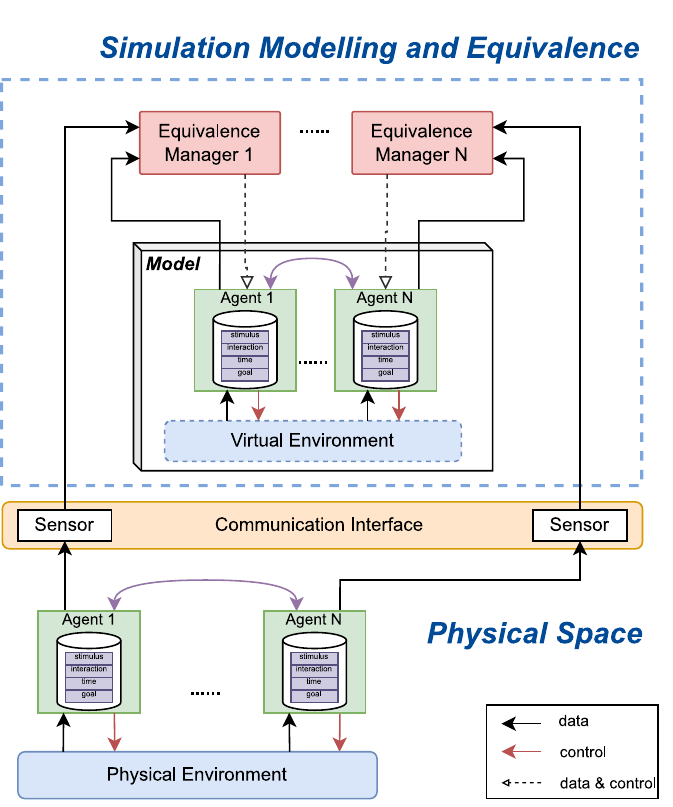}
  \captionof{figure}{Decentralisation of Equivalence Managers.}
  \label{fig: arch-decentralised}
\end{minipage}
\end{figure}

Since knowledge is divided into different levels in the agent's knowledge base, the equivalence manager should also maintain the equivalence of each dimension of fine-grained knowledge.
Fig. \ref{fig: arch-SA} shows the instantiation and refinement for the model equivalence of the generic reference model when the intelligence of the agents is modelled by self-awareness. 

The Equivalence Manager is composed of an equivalence checker and an updater. The equivalence checker (EC) contains a state equivalence checker, a knowledge equivalence checker, and an aggregator. 
The state EC checks equivalence by comparing the state transition of the physical world and the model.
The knowledge EC is further refined into four levels according to the levels of self-awareness: stimulus EC, interaction EC, time EC, and goal EC. 
Each fine-grained knowledge checker is responsible for the knowledge comparison of that level. For instance, the interaction EC monitors the interaction-related knowledge in the knowledge base of the physical agents and the virtual agents. When the interaction-related knowledge in the physical and virtual agents deviate from each other, the equivalence manager should update the interaction knowledge base of each virtual agent.
The aggregator coordinates the different ECs to decide which EC(s) to utilise. Such coordination may require a more sophisticated design and is out of the scope of this paper.
Finally, inequivalence detected by ECs will trigger the updater to update the model by simulation re-initialisation (on the states and/or knowledge) and/or model adaptation. 

The equivalence manager can be implemented as a central entity or decentralised to each virtual agent. 
In a centralised manner as shown in Fig. \ref{fig: arch-SA}, the manager checks the knowledge of all agents, and updates them at the same time. 
For the decentralisation, each virtual agent is assigned an equivalence manager as shown in Fig. \ref{fig: arch-decentralised}. Each equivalence manager only checks the knowledge and state of the agent that it is attached to. All equivalence managers also work asynchronously such that each manager decides by itself when to update its virtual agent. 
This paper particularly focuses on a centralised equivalence manager and the knowledge related to  interaction, i.e. equivalence checking by the interaction equivalence checker. The decentralisation is left as a future work.

\section{Knowledge Equivalence \label{sec:equivalence}}
Ensuring equivalence between the DT and the real world is essential for accurate simulation predictions and analysis. When ``twinning'' a system that learns knowledge, an additional requirement should be enforced on maintaining equivalence between the knowledge base of the physical agent and that of the corresponding virtual agent.
This section defines knowledge equivalence and presents different situations that lead to the knowledge deviation between the virtual system and the physical system.
The work presented in this paper focuses specifically on interaction-awareness. Though this awareness level deals with interactions, it backs itself up with both stimuli- and time-awareness knowledge in relation to these interactions. Knowledge can be implemented using different data structures.  
In this paper, the knowledge of an agent is represented as a weighted graph to model interaction with other agents.

\subsection{Defining Knowledge Equivalence}

First, consider a pair of physical and virtual agents with knowledge $k_i$ and $k_j$ respectively. The pair has identical  $\{A, P, K, \phi, \pi \}$ as modelled in Section \ref{sec:agent}. If the two agents are said to exhibit equivalence in their knowledge, then given the same perception, they should output and express the same action. 
That is, the two agents are  deemed to be  knowledge equivalent if and only if:
\begin{equation}
    \forall p \in P,\  \pi(p,k_i) = \pi(p,k_j) \label{eqn: equivalence}
\end{equation}
Notice that as mentioned in Section \ref{sec:agent}, the output of $\pi$ can either be one single action or a probability distribution of multiple actions. The comparison between two probability distributions can be quantified by well-established statistical techniques (e.g. histograms \cite{Elhabbash2016}).

For a multi-agent system and its model, the pair of agents of the virtual and physical worlds should strictly exhibit knowledge equivalent, if: 
for every agent $\alpha_i$ and its model $\alpha_i'$, equation (\ref{eqn: equivalence}) should hold.

However, exhaustively evaluating all possible perceptions is infeasible when the state space is too large or when the number of agents is also large. 
Additionally, it is equally time consuming to compare all possible perceptions at every wall clock time tick.

\subsection{Threats to Knowledge Equivalence} \label{sec:threats}

Assuming a correct and validated model by design, equivalence between the DT and the physical world can be threatened due to various reasons, thus causing deviation between the knowledge or/and states of the two intelligent systems.
Here we list below some of the major threats to equivalence.

\subsubsection{Temporal}
Temporal deviation refers to changes in the physical environment not being timely reflected in the DT model. This type of deviation can be attributed to the inadequate or suboptimal frequency of timely updates. The deviation is often noticed, when the time interval between two updates is long.

The updates can be infrequent and not timely for several reasons. 
Firstly, hardware bottleneck. The data processing capacity of the DT or network bandwidth may not, for example, sustain a high volume of incoming sensor data in a short period of time. 
Another reason is the need for minimising the energy consumption of the sensors. Frequent sensing and data transmission to the DT will cause the sensor may consume excessive energy, limiting its dependability and usage lifetime in the long run.
Thirdly, a sensor can be temporarily put into sleep mode when its data can be inferred by other sensors \cite{Minerva2020}, but at the risk of inaccurate estimation. 
Fourthly, the user (human or application) of the model may not require fine-grained high-frequency timely information.
Finally, sensors may fail to transmit data due to hardware failure.

The consequence of temporal deviation is that at any time step between two consecutive updates, the status of the physical world can be unknown to the DT. The DT may rely on data updated or estimates of previous runs, or may extrapolate its knowledge from other data sources.
To detect such a deviation, comparisons are needed at each updating time step.

\subsubsection{Unpredictable Environment Changes}
Deviation in knowledge can also happen due to unpredictable environmental changes that a running physical system may experience, and guards for triggering knowledge updates for these changes and surprises are missing in the design. 
One typical cause of unpredictability is partial observability -- the detail of the entire environment is not fully observed through sensors. Another cause is the model being designed with biased or simplified assumptions.
For example, the physical environment can be highly dynamic such that the environment model may not fully model all possible unforeseen scenarios, especially emergency events like car accidents in traffic-related applications, the appearance of sudden objects  obstructing the cars etc. 
Since agents interact with the environment, the deviation between the virtual and physical environment can cause the knowledge of virtual agents and physical agents to diverge, leading to different behaviours. Such a deviation will invalidate knowledge accumulated by the virtual system. To detect this type of deviation, continuous validation can be applied.

\subsubsection{Nondeterministic Agent Behaviours}
Intelligent self-aware agent may not act deterministically. If the physical agent and its virtual agent replica both act in a probabilistic manner, they are very likely to choose different actions even given the same perception and knowledge base. Different actions lead to different feedback from the environment. The two agents then may revise their knowledge bases differently given the different feedback. Different actions may also cause the virtual environment and physical environment to evolve in different trajectories. 
This type of deviation can be detected by directly comparing the actions made by the virtual agent and physical agent.

\subsubsection{Model Drift}
Similar to the discussion in \cite{Esfahani2013}, model drift refers to changes made in the DT not being reflected in the real world and vice versa. Note that model drift is not the same as the temporal deviation. The DT may modify the virtual system and synchronise the same changes in the physical system, but the synchronisation may fail, causing the two systems to drift apart. To give an example, the DT instructs a virtual robot to carry a parcel to a new location in the simulation, and then intends to synchronise the same action to the physical robot. However, during the operation of the physical robot, the parcel drops halfway during transportation. Without additional verification, the DT believes the parcel in the physical world has been transported successfully to the desired destination. The aggregation of this type of deviation over time will cause the model to drastically deviate from the physical world.
To detect this deviation, we need data from various additional sources to verify the synchronisation. In the above example, a camera can be installed on the robot to monitor the condition of the parcel as a means of verification. \newline

Anticipation of the above threats can be incorporated into the design of knowledge equivalence methods in order to mitigate possible deviation during runtime. In addition, these threats can be valid assumptions for the controlled experimental evaluation of DT system prototypes. This paper also uses combinations of threats to evaluate the knowledge equivalence-checking method proposed in the next section.


\section{A Methodology for Knowledge Equivalence Checking \label{sec:checking}}

The key to knowledge equivalence checking lies in the identification of potential knowledge drift through continuous online comparison.
In this regard, two fundamental questions arise: how to quantitatively identify knowledge drift, and when to update the simulation model.
The type of available sensor data dictates the possible continuous comparison approaches for online equivalence checking. Fig. \ref{fig: physical-virtual} shows the model used in this paper and it involves three types of available data that can be used in comparison: environment state, knowledge of agents, and actions of agents. 
While the comparison of the state is a straightforward approach, which is also reviewed in section \ref{sec:related work}, it does not consider the fact that physical agents maintain self-evolving knowledge bases. Therefore, two novel ways of comparison are proposed in this section based on the rest two types of data to directly check the equivalence of knowledge: knowledge comparison and action comparison.
A threshold-based tolerance approach is adopted to trigger model updates.

\subsection{Knowledge Comparison}
Knowledge comparison refers to directly quantifying the similarity of knowledge possessed by the virtual agent and physical agent.
The design of the similarity metric is specific to the type and representation of knowledge. 
Therefore, we specifically focus on agents that are interaction-aware.

\subsubsection{Knowledge Representation and Drift}

For systems where agents interact and collaborate with each other, a common practice is to represent interaction awareness as undirected weighted graphs \cite{Elhabbash2016,Esterle2014Socio-economicNetworks}. This paper also adopts graphs to represent interaction-related knowledge.
In the graph, each vertex represents an agent, and the edge weights are the knowledge of the interaction between agents. 
An example is to use weights to represent the collaboration history and tendency between two agents, which can be modelled by artificial pheromones \cite{Esterle2014Socio-economicNetworks}. For any agent $i$, if it has a high weight value with a certain agent $j$, then the weight means agent $i$ has been collaborating with $j$ recently in the past, and $i$ tends to collaborate more with $j$ in the future, rather than other agents with lower weight values.

Globally for the entire system with $m$ agents, there are at maximum \( l=\binom{m}{2}=m \times (m-1)/2 \) edges. We use $w_j$ to denote the weight value of the $j$th edge. The weights of all $l$ edges between physical agents at time $t$ are represented as a vector:
\begin{equation}
\mathbf{w}(t) = (w_1(t), w_2(t) ,w_3(t), ..., w_l(t))
\end{equation}
Similarly, the edge weights between virtual agents at the current time step $t$ are:
\begin{equation}
\mathbf{w}'(t) = (w_1'(t), w_2'(t), w_3'(t), ..., w_l'(t))
\end{equation}

Then to measure the knowledge drift of interaction-awareness between the simulation and physical world at time step $t$, we define a metric $drift()$ as the Euclidean distance between $\mathbf{w}(t)$ and $\mathbf{w}'(t)$:
\begin{equation}
    drift(\mathbf{w}(t), \mathbf{w}'(t)) 
             =|\mathbf{w}(t)-\mathbf{w}'(t)| 
             = \sqrt{\sum_{i=1}^{l} (w_i(t)-w_i'(t))^2} 
    \label{eqn: k-distance}
\end{equation}

\subsubsection{Workflow}
A threshold-based comparison workflow is designed to identify inequivalence of knowledge, which is shown in Algorithm \ref{alg: direct}. 
It is assumed that the simulation proceeds in real-time at the same pace as the wall clock time\footnote{The idea is to allow the simulation to run in parallel with the real world, such that at any moment, the simulation model always represents a snapshot of the real world, being neither ahead nor behind in time. Users would be able to inspect the current (estimated) status of the real world directly from the model, without waiting for the sensor data to be transmitted and collected. More importantly, such a real-time model is always ready to give predictions about how the real-world system would evolve starting from the current state of the world. }.
The algorithm continuously senses the knowledge bases in the simulated agents and physical agents at every wall clock time tick (lines 5-8). The time duration between two time ticks is specific to the application scenarios, and is out of the scope of this algorithm. 
The knowledge comparison is run every $q$ time ticks. A time window of length $l$ is used to calculate the cumulative knowledge drift within this window (lines 11-13), where the function $drift()$ is defined by Equation (\ref{eqn: k-distance}).
Then if the drift value is larger than a threshold $\theta$, the simulation is regarded as knowledge inequivalent to the physical world. The simulation then needs to be re-initialised with the latest world state and knowledge (lines 15, 16) as:
\begin{align}
    \begin{split}
        \mathbf{\sigma}'_t = \mathbf{\sigma}_t ,\ where\ \forall i\ \sigma'_{i,t} = \sigma_{i,t} 
    \end{split} \\
    \begin{split}
        \mathbf{w}'(t) = \mathbf{w}(t) ,\ where \ \forall i\ w'_i(t) = w_i(t)
    \end{split}
\end{align}

\begin{figure}
\centering
\scalebox{0.6}{
\begin{minipage}[t]{.8\textwidth}
    \removelatexerror
    \begin{algorithm}[H]
    \caption{Knowledge Comparison. \label{alg: direct}}
    \SetAlgoLined
    \KwIn{comparison time interval $q$, threshold $\theta$, comparison time window $l$}
    \KwData{History of knowledge of all physical agents $\mathbf{K}$, 
    history of knowledge of all virtual agents $\mathbf{K}'$, time $t$}
     $\mathbf{K}$ $\leftarrow$ []\;
     $\mathbf{K}'$ $\leftarrow$ []\;
     $t$ $\leftarrow$ 0\; 
     \While{ True }{
      $\mathbf{w}'(t) \leftarrow$ sense knowledge from simulation\;
      $\mathbf{w}(t) \leftarrow$ sense knowledge from real world\;
      $\mathbf{K}'[t] \leftarrow \mathbf{w}'(t)$\;
      $\mathbf{K}[t] \leftarrow \mathbf{w}(t)$ \;
        \tcp*[h]{Compare every $q$ time ticks} \\
        \If{t mod q == 0} { 
            $d \leftarrow 0$ \;
            \For{j = max(0, t-l) ... t}{
                $d \leftarrow d + drift(\mathbf{K}[j], \mathbf{K}'[j])$ \;
            }
            \If{d $>$ $\theta$ } {
                $\mathbf{\sigma}_t$ $\leftarrow$ physical world state at $t$ \;
                Re-initialise simulation with $\mathbf{\sigma}_t$, $\mathbf{K}[t]$\;
            }
        }
        wait for the next wall clock time tick\;
        $t$ $\leftarrow$ $t + 1$\;
     }
    \end{algorithm}
\end{minipage}
}
\hspace{0.2cm}
\scalebox{0.6}{
\begin{minipage}[t]{.8\textwidth}
    \removelatexerror
    \begin{algorithm}[H]
    \caption{Action Comparison. \label{alg: indirect}}
    \SetAlgoLined
    \KwIn{Comparison time interval $q$, threshold $\theta$, comparison time window $l$}
    \KwData{History of actions by all physical agents $\mathbf{A}$, 
    history of actions by all virtual agents $\mathbf{A}'$, time $t$}
      $\mathbf{A}$ $\leftarrow$ []\;
      $\mathbf{A}'$ $\leftarrow$ []\;
      $t$ $\leftarrow$ 0\; 
      \While{True}{ 
        $\mathbf{a}_t$ $\leftarrow $ sense actions from simulation\;
        $\mathbf{a}'_t$ $\leftarrow $ sense actions from real world\; 
        $\mathbf{A}[t] \leftarrow$ $\mathbf{a}_t$  \;
        $\mathbf{A}'[t] \leftarrow$ $\mathbf{a}'_t$ \; 
        \tcp*[h]{Compare every $q$ time ticks} \\
        \If{t mod q == 0} { 
            $d \leftarrow 0$ \;
            \For{j = max(0, t-l) ... t}{
                $d \leftarrow d + deviation(\mathbf{A}'[j], \mathbf{A}[j])$ \;
            }
            \If{d $>$ $\theta$ } {
                $\mathbf{k}_t$ $\leftarrow$ sense knowledge of all physical agents\;
                $\mathbf{\sigma}_t$  $\leftarrow$ sense physical world state \;
                Re-initialise simulation with $\mathbf{\sigma}_t$, $\mathbf{k}_t$ \;
            }
        }
        wait for the next wall clock time tick\;
        $t$ $\leftarrow$ $t + 1$\;
     }
    \end{algorithm}
\end{minipage}
}
\end{figure}

\subsection{Action Comparison}

Algorithm \ref{alg: direct} is only applicable when the drift of knowledge is clearly defined.
However, for more generalised situations where knowledge may not be represented as a graph or cannot be easily quantified, different approaches are needed. 
We propose an alternative approach that compares actions made by virtual and physical agents, since the action is the direct result of knowledge, and varies based on the level of knowledge the agent has.

\subsubsection{Action Deviation}
\label{sec:action_deviation}
In order to compare action, a metric that quantifies action deviation is proposed as follows.
The main idea is to measure how the actions made by all the virtual agents are different from the physical agents.

First of all, all the agents are assumed to have the same action set $A$ which contains a finite number of elements.
Then if the physical world is composed of $m$ agents in total, the deviation is calculated as

\begin{equation}
    deviation(\mathbf{a'}, \mathbf{a}) = \frac{1}{m} \sum_i^m \mathbb{1}_{a_i'= a_i}\ \ , where\ 
        \mathbb{1}_{a_i'= a_i} =
        \begin{cases}
          1 & a_i'= a_i\\
          0 & a_i'\ne a_i
        \end{cases}    
\end{equation}
where $a_i$ and $a_i'$ are the actions made by physical agent $i$ and its virtual counterpart $a_i'$.

\subsubsection{Workflow}
Algorithm \ref{alg: indirect} illustrates the workflow of action comparison, which is similar to knowledge comparison.
The main difference is in lines 5-8 and line 12, where the actions made by the physical agents and virtual agents are recorded and compared. In addition, instead of using $\mathbf{w}$ to denote knowledge specifically represented as a graph, we use $\mathbf{k}$ to represent knowledge in order to be generic.

\section{Evaluation} \label{sec:evaluation}

This section evaluates the effectiveness of the previously proposed two knowledge equivalence checking methods in maintaining an equivalent DT. The evaluation is conducted through a case study using smart mobile cameras.
Maintaining knowledge equivalence involves two repetitive steps: checking for knowledge discrepancies and updating the knowledge. 
As mentioned in sections \ref{sec:equivalence} and \ref{sec:checking}, the main premise of the paper is that 
1) maintaining knowledge equivalence is crucial for maintaining an equivalent DT model that replicates networked intelligent self-aware systems, and that
2) the methods used for checking knowledge equivalence are efficient and have low overhead in keeping the DT model up-to-date.
Two factors are considered in the evaluation: 1) updating the model is resource-intensive, since it requires re-initialisation of all variables, which is time-consuming, and 2) comparison takes time and memory space.

The evaluation will utilise the following metrics:

\begin{itemize}
    \item \textit{Number of updates}: this metric corresponds to the total number of updates within a given time period. This metric is indicative of potential overheads that can be incurred as a result of the updates, where more updates may require frequent re-configuration and re-initialisation of the simulation. Once knowledge in-equivalence is identified, an update will follow; henceforth, the number of updates is equal to that of the number of in-equivalence checks.
    \item \textit{Average utility deviation}: This metric relates to the difference between the task goal satisfaction (defined by a utility function) observed in the real system and the simulated modelled system. This metric indicates simulation validity, averaged over the time of observations. 
    The utility function is application-specific, but the metric of utility deviation within $n$ consecutive time units can be defined as: 
    \begin{equation}
        \text{Avg. utility deviation} = \frac{1}{n} \sum_{t=1}^{n} | u(t) - u'(t) | \\
    \end{equation}
    where $u(t)$ is the utility of the real system at time $t$ and $u'(t)$ is the utility of the modelled system at time $t$.
\end{itemize}

The minimisation of both metrics is essentially conflicting in objectives. When the model is updated more frequently, the simulation of the modelled system strives for fidelity with the real system. Conversely, with less frequent updates, the simulation can easily drift away from the real system. 
To address this problem, we assess each of our proposed checking methods by the \textit{Pareto efficiency} of the method's solution set. 
The solution set is obtained by various possible configurations of that method against the two objectives.
Pareto efficiency is based on the \textit{non-domination} relation between solutions. A solution point $x$ dominates $y$ if and only if $x$ is at least as good as $y$ in both objectives and better in at least one \cite{While2006Hypervolume}. A point $x$ is \textit{non-dominated} if and only if no other points dominate $x$. The set of all mutually non-dominated solutions is called the \textit{Pareto front}.
The experimental analysis aims to answer the following two questions:

\renewcommand{\theenumi}{\textbf{Q\arabic{enumi}}}

\begin{enumerate}
    \item \label{Q_update} Does knowledge update lead to a high fidelity replica, such that the simulation model closely describes the real world and exhibits smaller utility deviation?
    \item \label{Q_checking} Does knowledge equivalence checking achieve more Pareto efficient results than state equivalence checking under different uncertainties?
\end{enumerate}

\subsection{An Illustrative Example}

The evaluation adopts a modified version of the real-time tracking by distributed smart mobile cameras\footnote{All elements used for the experimental analysis presented in this paper including the source code, parameters, data and results are available at \url{http://digitwins.github.io}} \cite{Esterle2017}.
This example is one of the various studies where self-awareness has been applied and has demonstrated the advances in achieving self-organising behaviours \cite{rinner_self_2015}.
In addition, the study of distributed smart cameras itself has also been actively studied regarding various decentralised approaches and different variations \cite{Esterle2014Socio-economicNetworks,yang_distributed_2022,pianini_collective_2022}.
Our example based on \cite{Esterle2017} represents a typical setup for how networked self-aware agents in a spatial environment can self-learn by accumulating knowledge of interaction and self-organise to finish specific tasks.
The example also fits this paper's assumption of having decentralised agents that self-evolve their knowledge bases with interaction-awareness.
This example case study is complex by exhibiting emergent properties: minor differences in the state will easily affect the behaviour of agents and diverge the self-organising behaviours of other networked agents thereafter.

\begin{figure}[!t]
\centering
\includegraphics[width=0.6\linewidth]{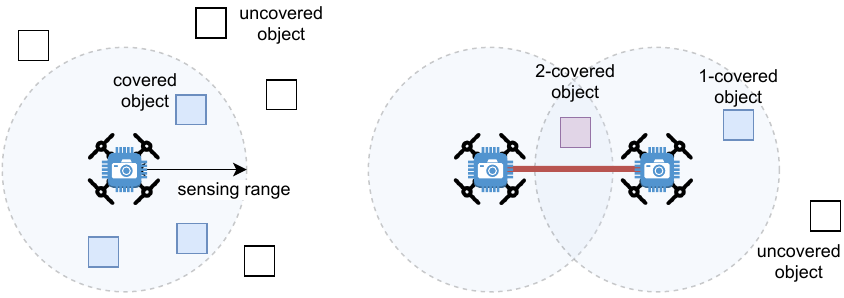}
\caption{k-coverage from the bird's-eye view.}
\label{fig: example}
\end{figure}

In the example, a fixed number of objects and intelligent drones move in a bounded 2-D space. The drones are equipped with onboard cameras to collaboratively track and monitor objects for surveillance purposes. Each camera is assumed to have an omnidirectional (360-degree) view with a fixed sensing range.
Each drone is controlled by an embedded autonomous self-aware agent which decides its actions. 
Each object moves towards a fixed direction at a constant speed, and will bounce back when it hits the boundary of the 2-D space. Objects can become important or unimportant spontaneously. 
The goal of the system of drones is to maximise \textit{k-coverage}: trying to ensure every \textit{important} object is monitored by at least $k$ cameras.
We further assume that as long as an object is located within the sensing range of any camera, the object is regarded as covered by that camera. An object is \textit{k-covered} if it is covered by at least $k$ cameras at the same time (see Fig. \ref{fig: example}).

\subsubsection{Utility of the Task Goal}

To measure the satisfaction of task goal (\textit{k-coverage}) over time, a \textit{utility function} is used \cite{Esterle2017}. Suppose there are $N_{all}$ objects in the environment. Let $N_{k}(t)$ be the number of objects being k-covered at time $t$. Then the utility at moment $t$ is defined as the ratio of k-covered objects: 
\begin{math}
u(t)=\frac{N_{k}(t)}{N_{all}}.
\end{math}

\subsubsection{Self-Aware Agent}

Since no single agent has a global view of the world (due to limited sensing capability), each agent is assumed to collaborate with others based on message communication. 
An agent can request other agents to help track the same object by advertising the object ID and position. Other agents can then decide whether to accept the request and move towards the advertised object.
To avoid broadcasting to all other agents which is costly in communication, each agent is assumed \textbf{interaction-aware} to intelligently decide which ones to send messages to and which request messages to accept.

First of all, stimulus-awareness is the basis of self-awareness. The stimuli in our example are the local perception of the agent, which includes the received messages, the current position of itself, as well as the current positions and the importance of all objects within its camera range. An agent is able to use the stimuli to decide which direction to move to track a certain object.

Beyond stimuli, the knowledge of interaction in this example is the status of co-covering. 
A drone is able to identify the coverage of other drones of objects as itself in a given time; this can be achieved through short-range wireless sensing technology as assumed in the original example \cite{Esterle2017}.
Such knowledge of interaction is modelled as artificial pheromones, represented as a weighted graph, and maintained by each agent in its local memory.
Agents are vertices in the graph. The edge weights reflect the recency of co-covering collaboration.
If any two agents are both covering one or more same important objects, then the weight of the edge connecting the two agents will be ``strengthened'' by adding a constant value $\delta$.
All edge weights will also ``evaporate'' through time by multiplying a discount factor $\gamma$ at the end of every time unit \cite{Esterle2017,Esterle2014Socio-economicNetworks}.

With the knowledge of interaction, an agent tends to communicate with agents that it has collaborated with very often recently. When requesting help, the agent shall only notify the $(k+1)$ agents with which it has the strongest weights. When deciding which request to accept, the agent will rank the received messages first by edge weight and then by the received time. The agent will only accept the most recent request sent by the agent that has the highest edge weight.
A more detailed description of the behaviour of a single agent is shown in Algorithm \ref{alg: camera} in appendix \ref{Agent behaviour in the motivating Example} which is based on \cite{Esterle2017}. Each agent in the system is designed as Algorithm \ref{alg: camera}.


\subsection{Prototype Implementation}

We use two simulators to implement the example and its DT as shown in Fig. \ref{fig: implement}.
The two simulators run in parallel, denoted as \textit{Simulator \textbf{P}} and \textit{Simulator \textbf{D}} to represent the \textbf{P}hysical world and its \textbf{D}igital twin replication, respectively.
Both simulators are implemented by the agent-based simulation engine Repast Simphony \cite{North2013ComplexSimphony}, and can simulate the behaviours of objects and drones. The two simulators are identical \texttt{jar} executables, but multiple types of uncertainties are introduced between the two, which is explained later in section \ref{sec: secnarios}. 
The simulator initialises its starting scene by loading an XML file, which contains the information of all drones and objects as shown in Table \ref{tab: data}. The simulation state $\sigma_t$ as defined by Equation \ref{eqn: state}  is composed of the positions (x-y coordinates) of all objects and drones at time $t$.

The equivalence manager is implemented in Python and uses Py4J\footnote{https://www.py4j.org/} to interact with the two simulators.
It is able to sense data listed in Table \ref{tab: data} from both simulators at every time step.
When an update of the DT is needed, the equivalence manager re-initialises \textit{Simulator D} by passing all data in Table \ref{tab: data} sensed from \textit{Simulator P} as an XML file.

\begin{minipage}[t]{0.3\textwidth}
    \centering
    \captionsetup{type=table}
    \captionof{table}{Types of data accessible in simulation.}\label{tab: data}
    \begin{tabular}[t]{cl}
        \toprule
        \textbf{Entity} & \textbf{Attribute Data} \\
        \midrule
        Drone & ID \\
            & position \\
            & received messages \\
            & interaction graph \\
        \hline
        Object & ID \\
            & position \\
            & direction \\
            & importance \\
        \bottomrule
    \end{tabular}
\end{minipage}
\hfill
\begin{minipage}[t]{0.65\textwidth}
    \raggedleft
    \captionsetup{type=figure}
    \includegraphics[width=0.9\linewidth]{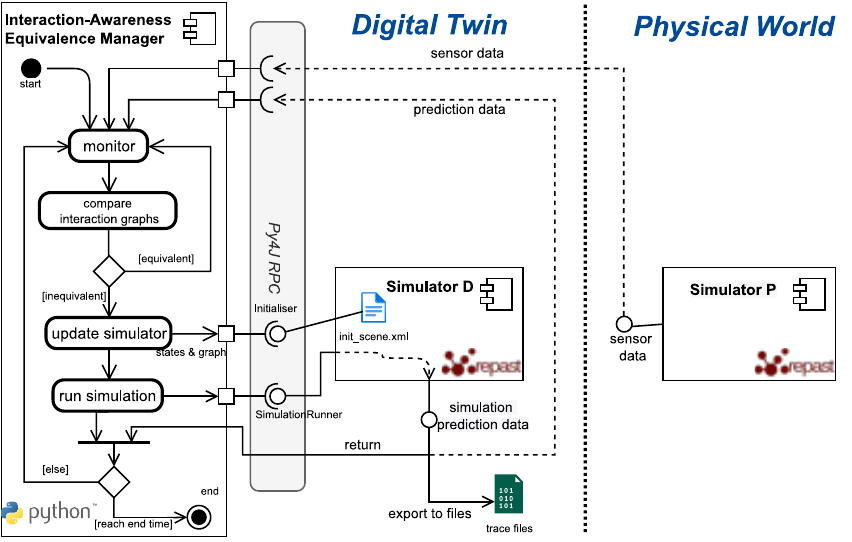}
    \captionof{figure}{The implementation architecture for the prototype.}\label{fig: implement}
\end{minipage}

\subsection{Experimental Frame}

\subsubsection{Evaluation Scenarios \label{sec: secnarios}}

\begin{figure*}[hb]
\centering
    \begin{subfigure}[b]{0.16\textwidth}
         \centering
         \includegraphics[width=\linewidth]{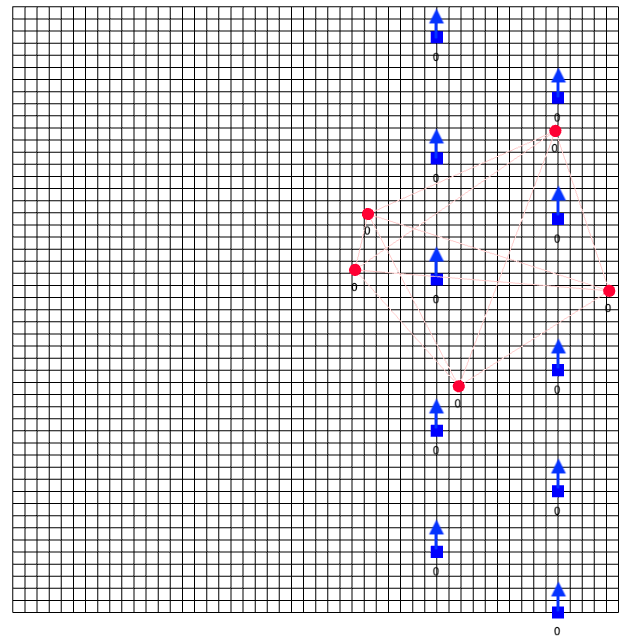}
         \caption{Scene 1: 10 objects, 5 drones.}
         \label{fig: scene1}
     \end{subfigure}
     \hfill
     \begin{subfigure}[b]{0.16\textwidth}
         \centering
         \includegraphics[width=\linewidth]{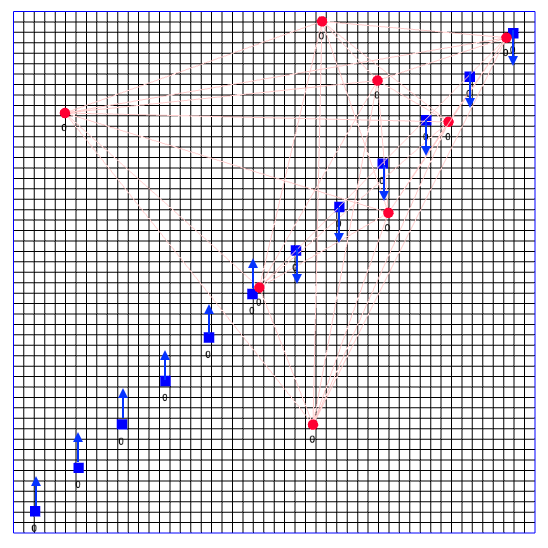}
         \caption{Scene 2: 12 objects, 8 drones.}
         \label{fig: scene2}
     \end{subfigure}
     \hfill
     \begin{subfigure}[b]{0.16\textwidth}
         \centering
         \includegraphics[width=\linewidth]{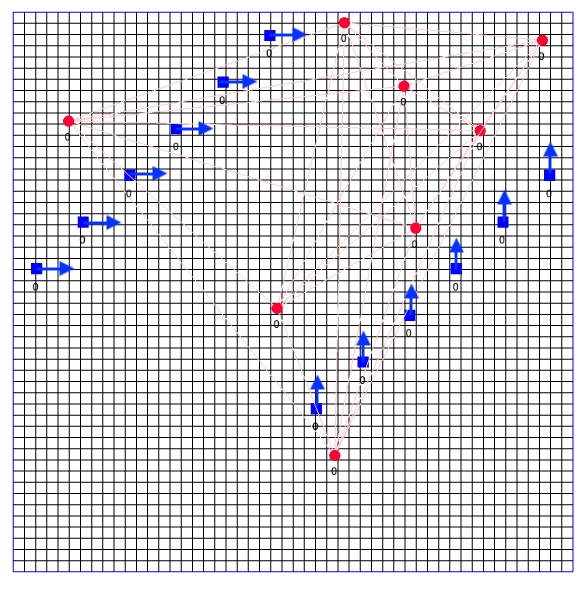}
         \caption{Scene 3: 12 objects, 8 drones.}
         \label{fig: scene3}
     \end{subfigure}
     \hfill
     \begin{subfigure}[b]{0.16\textwidth}
         \centering
         \includegraphics[width=\linewidth]{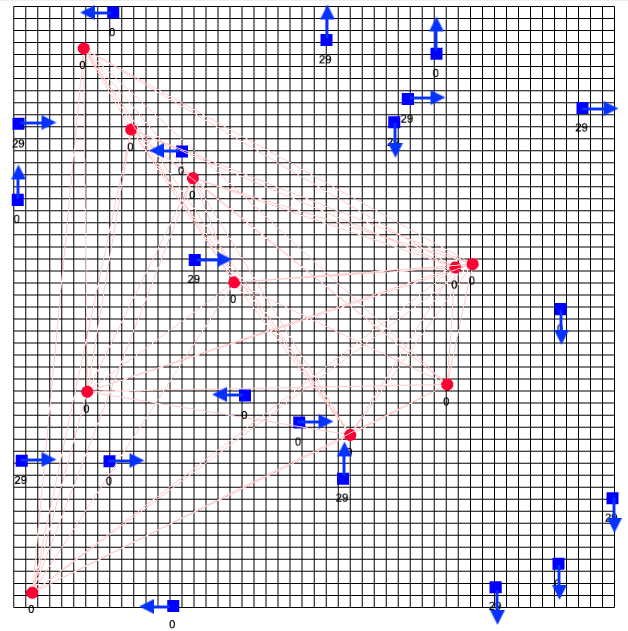}
         \caption{Scene 4: 20 objects, 10 drones.}
         \label{fig: scene4}
     \end{subfigure}
     \hfill
     \begin{subfigure}[b]{0.16\textwidth}
         \centering
         \includegraphics[width=\linewidth]{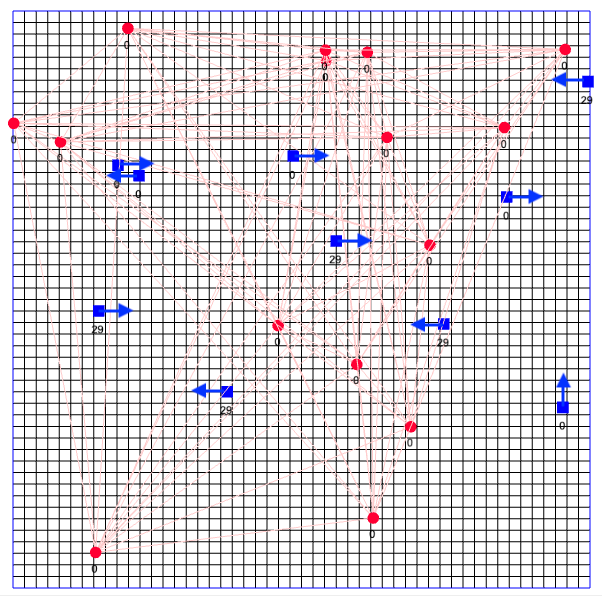}
         \caption{Scene 5: 10 objects, 15 drones.}
         \label{fig: scene5}
     \end{subfigure}
     \hfill
     \begin{subfigure}[b]{0.16\textwidth}
         \centering
         \includegraphics[width=\linewidth]{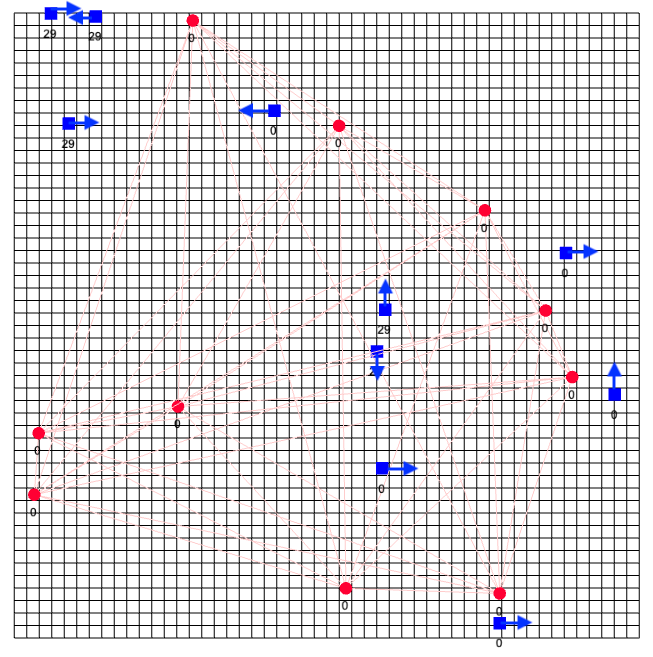}
         \caption{Scene 6: 10 objects, 10 drones.}
         \label{fig: scene6}
     \end{subfigure}
\caption{The initial snapshot of the evaluation scenes.
We consider a 2-D world of size 50 * 50. 
The blue rectangles are the objects. The red circles are the drones. 
Scenes 4, 5 and 6 are generated by randomly positioning the objects and drones and randomly assigning the initial moving direction of each object to be one of the four: east, north, west and south.
}
\label{fig: scenario}
\end{figure*}

To emulate the situations in real-world settings, biases and uncertainties are considered. Following on the discussions of section \ref{sec:threats}, we consider the following three threats to equivalence:
\renewcommand{\theenumi}{T\arabic{enumi}}

\begin{enumerate}
    \item \label{T_param} Unpredictable environment changes: biased model parameters.
    \item \label{T_env} Unpredictable environment changes: the environment is partially observable; data out of the sensor range are unknown. 
    \item \label{T_rand} Nondeterministic agent behaviours.
    
\end{enumerate}

First, for \ref{T_param}, systematic deviation of an object could be encountered in the physical environment, but may not be considered in the model of the object. 
In the experiment, the simulation model assumes its objects to have experienced systematic deviation in movement, where 
each physical object moves with a systematic error angle of 3 degrees to its left, but the model is not aware of such a deviation.
Such an error can be common. For instance, if the object moves with wheels, then the error could be caused by the manufacturing imprecision of the wheels.

Second, for \ref{T_env}, if the DT is only able to acquire data from the drones in the physical world, then the positions of objects that are not covered by any drone are deemed to be unknown. Therefore, the behaviour of the entire system cannot be simulated accurately. These unknown positions can be estimated (e.g. through past trajectories of objects), but the estimation techniques are out of the scope of this paper. In the experiments, we assume the DT can estimate these positions with a random error $\epsilon$ on both $x$ and $y$ coordinates.

Third, for \ref{T_rand}, the virtual agents and physical agents may exhibit different behaviours. When randomly deciding which object to track at line 11 of Algorithm \ref{alg: camera}, the physical agent and its virtual counterpart may take different choices. 
In the experiments, the virtual agents and physical agents are provided with different random seeds.

We construct two conditions combining two threats, where each of the conditions, I and condition II are applied to six different starting scenes (shown in Fig. \ref{fig: scenario}) respectively: (a) 
Condition I:  \ref{T_param} and \ref{T_rand} (b) Condition II: \ref{T_param} and \ref{T_env}.
Therefore, in total $2 \times 6 = 12$ different evaluation scenarios are designed. Each scenario is denoted by its condition and initial scene. For instance, scenario I-3 refers to the combination of Condition I and scene 3.

\subsubsection{Simulation Setup \label{sec: simulation_setup}}

The simulation uses discrete logical time. 
The model parameters for the two simulators are set as follows. 
The environment is designed to be a 50 $\times$ 50 2-D world with boundaries. The task goal is to maximise 2-coverage, with $k=2$. 
Each object moves 1 unit distance every time step and becomes important or unimportant every 30 time steps.  
The sensing range of each camera is 10. The parameters of edge weights are $\gamma=0.9$ and  $\delta=1$. 
The random error $\epsilon$ of position estimation as described in section \ref{sec: secnarios} is uniformly sampled from [-5, 5]. This is a relatively large error since the uncertainty spans 20\% of the side length of the 2D world, and it is at the same scale as the sensing range of the camera. Such an error assumes poor estimation of the unobserved environment by the DT.
In the beginning, the two simulators are initialised with identical states and zero knowledge.
Each experiment configuration is run for 1000 simulation time steps and repeated 5 times if not specified otherwise.

\subsection{Simulation Validity of Knowledge Updates}

Firstly, the performance of knowledge updates is evaluated.
The aim is to investigate how different knowledge update strategies, when applied at different time steps of the simulation, can affect utility deviation over time. The evaluation is intended to investigate the question \ref{Q_update} for the necessity of knowledge synchronisation.
In particular, three update strategies used for synchronising the DT are evaluated, each differs in how the knowledge is updated:
\begin{itemize}
    \item \textit{Keep knowledge}: update all data items listed in Table \ref{tab: data}, except for the interaction graph, and keep the graph in the simulated system unchanged.
    \item \textit{Clear knowledge}: update all data items in Table \ref{tab: data}, except for the interaction graph, and set all edge weights to zero.
    \item \textit{Update knowledge}: update all data items listed in Table \ref{tab: data}
\end{itemize}

\textit{Keep knowledge} and \textit{Clear knowledge} are two baselines. They represent two setups where the model update does not pay attention to the knowledge. Then the knowledge either remains unchanged or is reset. Only state variables and essential attributes of agents are synchronised to the DT.  Instead, \textit{update knowledge} synchronises not only the latest state but knowledge to the DT.

Each experiment is designed to run for 100 time steps, which contains two phases: \textit{knowledge accumulation} phase and \textit{evaluation} phase.
Each phase lasts for 50 time steps. The beginning of \textit{knowledge accumulation} is called the \textit{start time}, and the end is called the \textit{update time}.
At \textit{start time}, the simulated system is initialised by replicating the snapshot of the real system, which contains all data listed in Table \ref{tab: data}. At the \textit{knowledge accumulation} phase, both systems run for 50 time steps to allow their knowledge to evolve. The two systems also gradually deviate from each other because of condition I. Next, at \textit{update time}, the three update strategies are applied respectively to the simulated system by sensing the real system. Finally, at the \textit{evaluation} phase, both systems run another 50 time steps, and the utility deviation is evaluated.

To emulate updating at different points in time, for each of the scenarios I-1 to I-6, 30 time stamps are randomly sampled from 1 to 900. Each time stamp serves as the \textit{start time} for one experiment. Each experiment is repeated 10 times. 

\begin{figure}
\centering
\begin{minipage}[b]{.5\textwidth}
  \centering
  \includegraphics[width=0.8\linewidth]{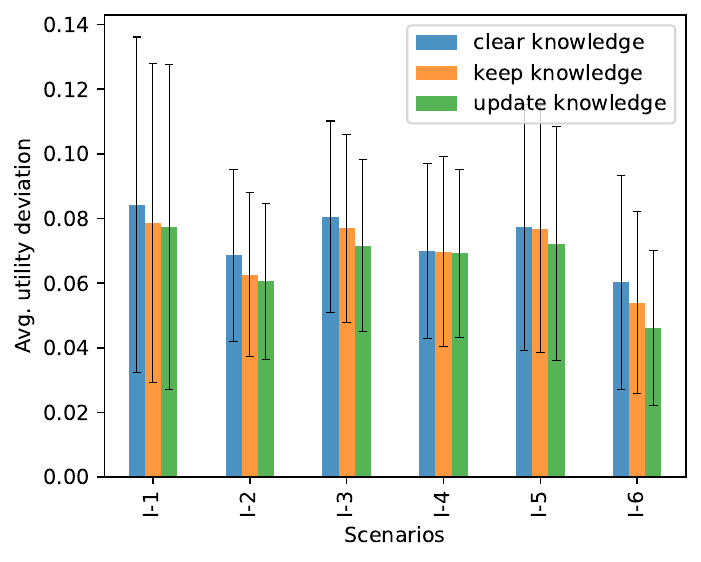}
  \captionof{figure}{Knowledge update strategies in 6 scenarios.}
  \label{fig: knowledge update}
\end{minipage}%
\begin{minipage}[b]{.5\textwidth}
  \centering
  \includegraphics[width=\linewidth]{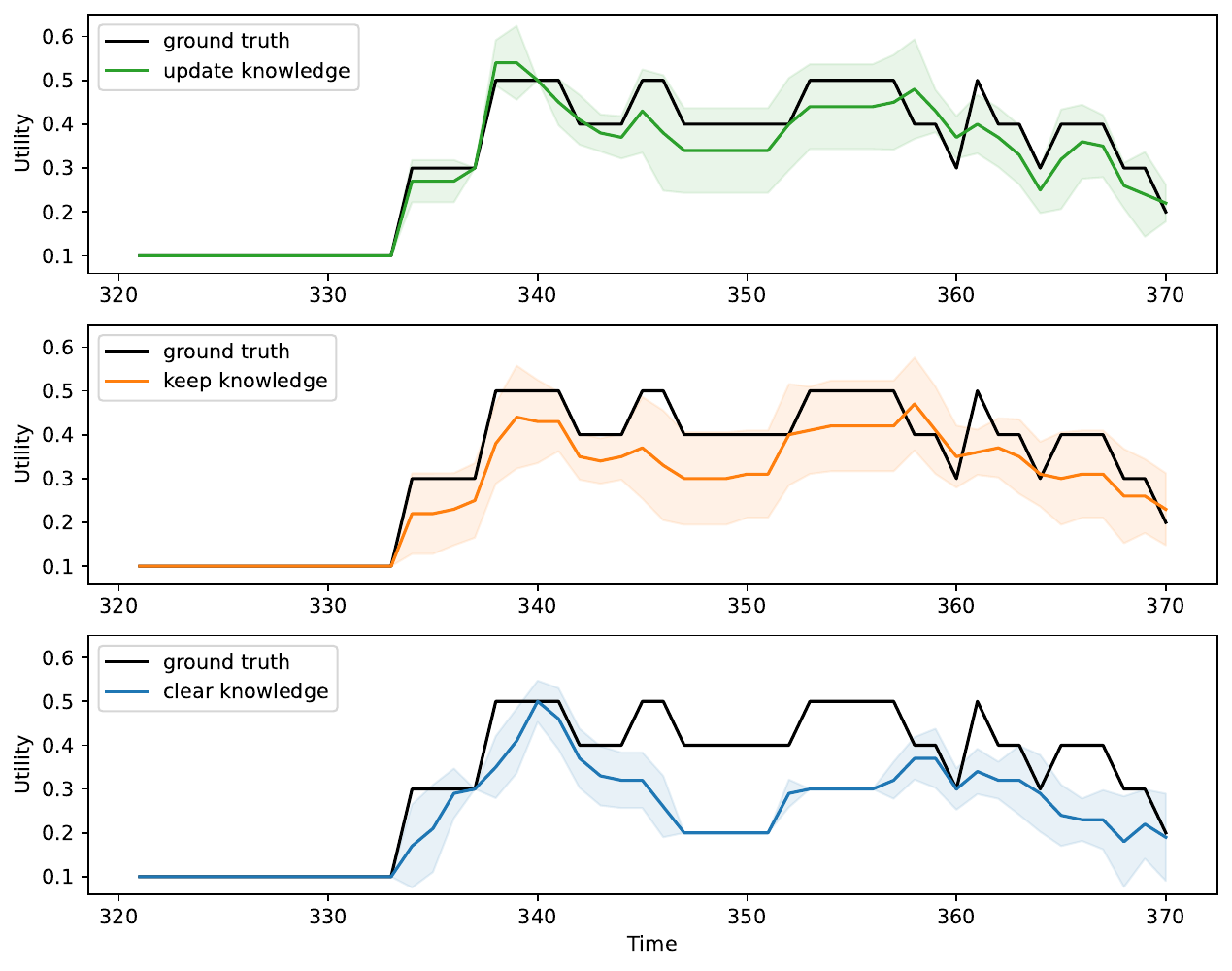}
  \captionof{figure}{2-coverage utility of three knowledge update strategies, after updating at time 320 in Scenario I-6.}
  \label{fig: knowledge update lines}
\end{minipage}
\end{figure}

The average results for $30\times 10 = 300$ runs of each scenario are shown in Fig. \ref{fig: knowledge update}. 
The deviation of each bar is relatively large because even for one certain scenario, the system behaviours during different time periods are different. Then the performance of utility deviation may fluctuate among these 30 randomly sampled time periods.
Nevertheless, by taking the average, the figure shows that for all the scenarios, the \textit{update knowledge} strategy performs the best by achieving the lowest utility deviation. 
If knowledge is kept the same as before, the old knowledge can still take effect but the accuracy of utility is lower. 
If knowledge is reset at the update time, the simulation generally cannot guarantee the accuracy of utility as the other two strategies.
Since different knowledge leads to different actions of the agent, the resultant task utility may also be affected. Therefore, when updating, in addition to maintaining the same state in the simulated system, the same knowledge should also be maintained.

In addition, Fig. \ref{fig: knowledge update lines} shows the utility of scenario I-6 during the \textit{evaluation phase}. The start time and update time are set as 270 and 320, respectively. The figure clearly shows that updating knowledge ensures the simulation most accurately ``follow'' the ground truth utility.
Notice that the simulation starts to deviate only after time 333 in both three sub-figures.
This is first because all objects switch their status of importance at time 330 (as assumed in Section \ref{sec: simulation_setup}). The important objects covered by drones then become unimportant. Therefore, the drones need to search and cover other objects that just became important, which leads to different movement patterns than patterns before time 330.  
Also, Condition I causes the positions of objects in the simulation to deviate from the real objects. 
With different drone movement patterns and different object positions, the utility deviates after time 333.

\subsection{Pareto Efficiency of Knowledge Equivalence Checking \label{sec: pareto evaluation}}

Next, we evaluate our proposed knowledge equivalence checking methods: knowledge comparison and action comparison, which do not assume a fixed update frequency. When updating Simulator D, all data in Table \ref{tab: data} (state and knowledge) are replicated from Simulator P. This evaluation is intended to study the question \ref{Q_checking}.

We regard \textit{state comparison} as the baseline, which is shown in Algorithm \ref{alg: state}. 
A state is composed of the positions of all the cameras and objects. The deviation of two states is also measured by their Euclidean distance as:
\begin{equation}
deviation(\mathbf{\sigma}_t, \mathbf{\sigma}'_t) 
             =|\mathbf{\sigma}_t-\mathbf{\sigma}'_t| 
             = \sqrt{\sum_{i=1}^{l} (\sigma_{i,t}-\sigma'_{i,t})^2}
\end{equation}

\begin{figure}
\centering
\scalebox{0.6}{
\begin{minipage}[t]{0.8\textwidth}
    \removelatexerror
    \begin{algorithm}[H]
    \caption{State Comparison. \label{alg: state}}
    \SetAlgoLined
    \KwIn{Comparison time interval $q$, threshold $\theta$, comparison time window $l$}
    \KwData{History of physical world state $\mathbf{\Sigma}$, 
    history of virtual world state $\mathbf{\Sigma}'$, time $t$}
      $\mathbf{\Sigma}$ $\leftarrow$ []\;
      $\mathbf{\Sigma}'$ $\leftarrow$ []\;
      $t$ $\leftarrow$ 0\; 
      \While{True}{
        $\mathbf{\sigma}_t \leftarrow $ sense from simulation\;
        $\mathbf{\sigma}'_t \leftarrow $ sense from real world\; 
        $\mathbf{\Sigma}[t] \leftarrow \mathbf{\sigma}_t$ \;
        $\mathbf{\Sigma}'[t] \leftarrow \mathbf{\sigma}'_t$\; 
        \tcp*[h]{Compare every $q$ time ticks} \\
        \If{t mod q == 0} { 
            $d \leftarrow 0$ \;
            \For{j = max(0, t-l) ... t}{
                $d \leftarrow d + deviation(\mathbf{\Sigma}'[j], \mathbf{\Sigma}[j])$ \;
            }
            \If{d $>$ $\theta$ } {
                $\mathbf{k}_t$ $\leftarrow$ sense knowledge of all physical agents\;
                Re-initialise simulation with $\mathbf{\sigma}_t, \mathbf{k}_t$ \;
            }
        }
        wait for the next wall clock time tick\;
        $t$ $\leftarrow$ $t + 1$\;
     }
    \end{algorithm}
\end{minipage}
}
\hspace{0.2cm}
\scalebox{0.6}{
\begin{minipage}[t]{.8\textwidth}
    \removelatexerror
    \begin{algorithm}[H]
    \caption{Fine-Grained Action Deviation.\label{alg: action2}}
    \SetAlgoVlined
    \KwIn{The action made by a physical agent, the action made by the virtual agent}
    \KwOut{The deviation of two actions}
      \lIf{both actions are \texttt{random walk}} {\Return 0}
      \If{both actions are \texttt{follow}} {
        \uIf{both agents follow the same object} {
          \Return 0\;
        }
        \lElse{\Return 1}
      }
      \If{both actions are \texttt{respond-and-follow}} {
        $d \leftarrow 0.4$\;
        \If{both agents respond to the same agent} {
          $d \leftarrow d - 0.2$\;
        }
        \If{both agents follow the same object} {
          $d \leftarrow d - 0.2$\;
        }
        \Return $d$\;
      }
      \If{both actions are \texttt{notify-and-follow}} {
        $c \leftarrow $ the number of commonly notified agents by the two given agents\;
        \uIf{both agents follow the same object} {
          \Return $0.5 \cdot (1-\frac{c}{k-1})$\;
        }
        \lElse{\Return $0.5 \cdot (1-\frac{c}{k-1}) + 0.5$}
      }
      \If{one action is \texttt{follow} and the other is \texttt{notify-and-follow}} {
        \uIf{both agents follow the same object} {
          \Return 0.5\;
        }
        \lElse{\Return 1}
      }
      \Return 1\;
    \end{algorithm}
\end{minipage}
}
\end{figure}

There are three parameters for each comparison method. We assume the comparison is made every time step $q=1$ and the time window is $l=1$. 
The performance of the comparison method is largely dependent on the threshold value $\theta$, which defines the tolerance to deviation. We evaluate various different values of $\theta$ for each method, and contrast the three methods by their Pareto efficiency. Experiments for each threshold value are repeated 5 times.

Knowledge comparison and state comparison are both applied to each of the 12 scenarios.
Action comparison is applied to scenarios II-1 to II-6.
In action comparison, we view the action at a coarse grain and only distinguish 4 different actions: \texttt{follow}, \texttt{notify-and-follow}, \texttt{respond-and-follow}, and \texttt{random walk}.

\subsubsection{Performance Trade-off Due to Threshold Values}
\begin{figure}[!t]
\centering
\includegraphics[width=\linewidth]{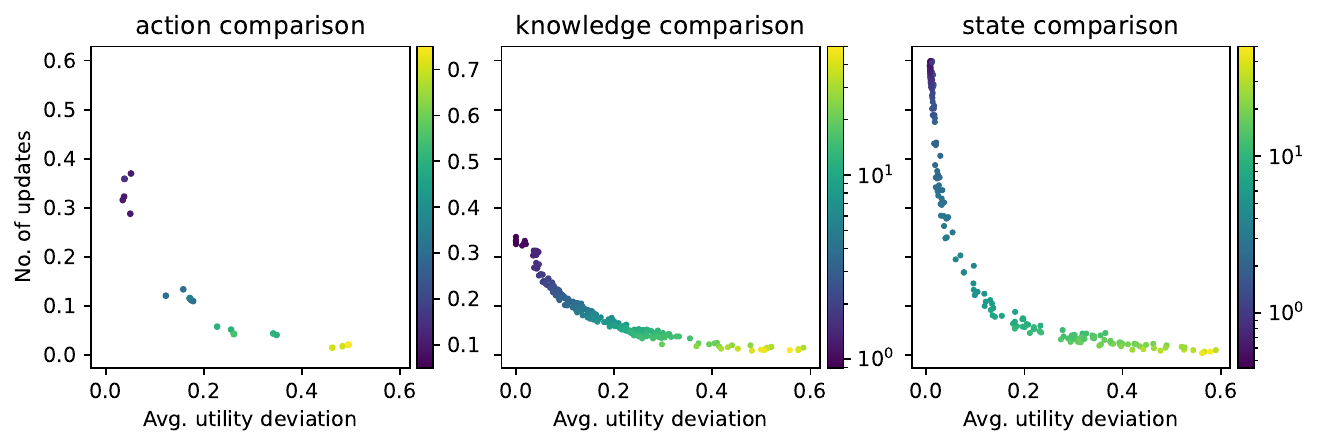}
\caption{Solutions (after normalisation) obtained by different threshold values in scenario I-1. Values of the threshold are shown in different colors. }
\label{fig_threshold}
\end{figure}

The different solutions obtained by various threshold values can be seen in Fig. \ref{fig_threshold}, in which we apply the three comparison methods to scenario I-1. 
The trade-off between the two metrics can be observed for each of the methods.
When the threshold value is small, the number of updates is large and the utility deviation is small. 
Conversely, a larger threshold will lead to less number of updates but a larger utility deviation.
This result validates the existence of a trade-off curve between the two metrics.
Therefore, Pareto efficiency is then used to compare the solution sets of the three methods. 

\subsubsection{Analysis by Pareto Efficiency}
\begin{figure*}[!t]
\centering
\includegraphics[width=\linewidth]{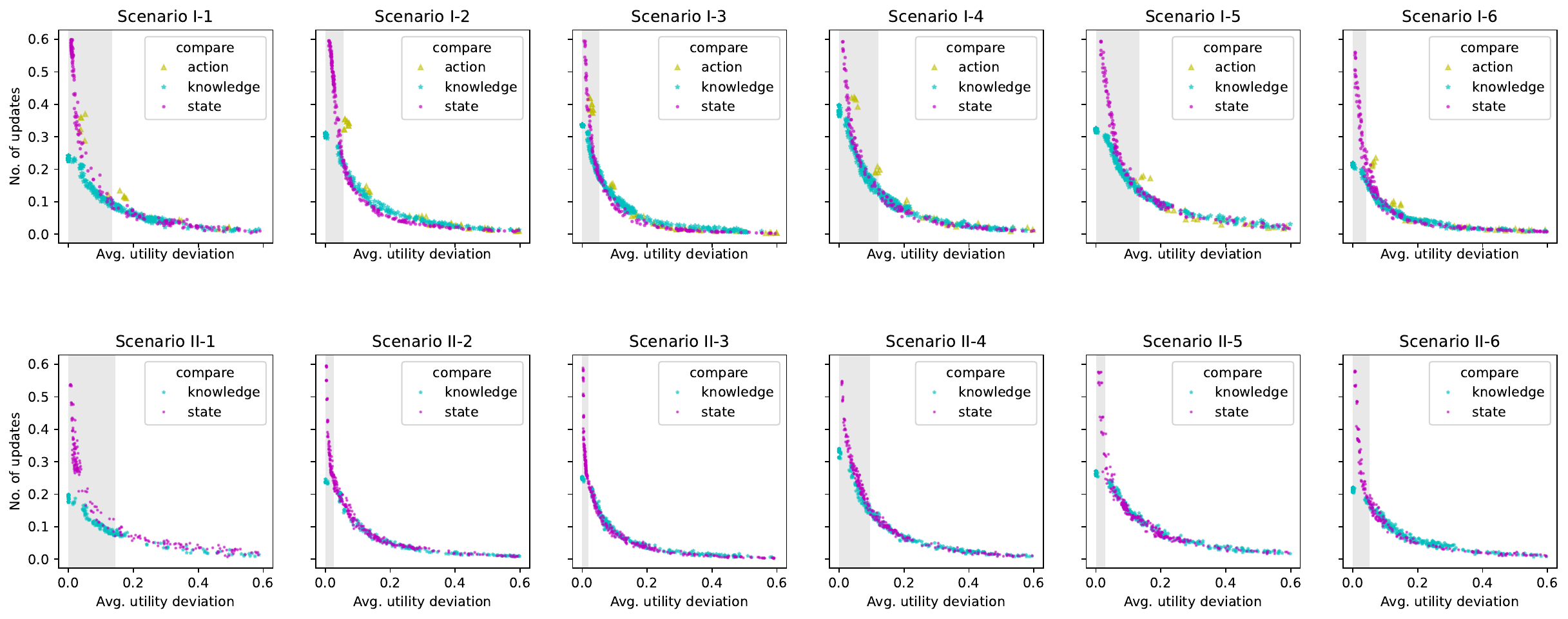}
\caption{Simulation results for all 12 scenarios. The x-axis is normalised by the value of the average utility deviation obtained when no update is involved. The values used for normalisation are (row by row, from left to right for each row in the figure): 0.1436, 0.1452, 0.2224, 0.1343, 0.1165, 0.1416, 0.1519, 0.1519, 0.2167, 0.1449, 0.1250, 0.1475. The y-axis is normalised by 1000, which is the worst situation where the simulation should be updated each time unit. Within the highlighted area in grey, all solutions of state comparison are dominated by knowledge comparison.}
\label{fig_sim}
\end{figure*}

The results for the experiments on all 12 scenarios are shown in Fig. \ref{fig_sim}. 
In all the 12 scenarios, knowledge comparison shows more Pareto efficient results when the average utility deviation is small, which are shown in the highlighted area of each sub-figure.
A clear divergence trend can be observed in the solutions that as the average utility deviation decreases, solutions of knowledge comparison start to dominate state comparison. Knowledge comparison also exhibits a flatter curve in contrast to state comparison.
In the extreme case, knowledge comparison achieves 0 utility deviation with much fewer updates than state comparison. 
However, when the utility deviation is larger, state comparison can be sometimes better or similar to knowledge comparison.
In addition, action comparison is generally slightly worse than both other methods.

\begin{figure}[!t]
\centering
\includegraphics[width=0.7\linewidth]{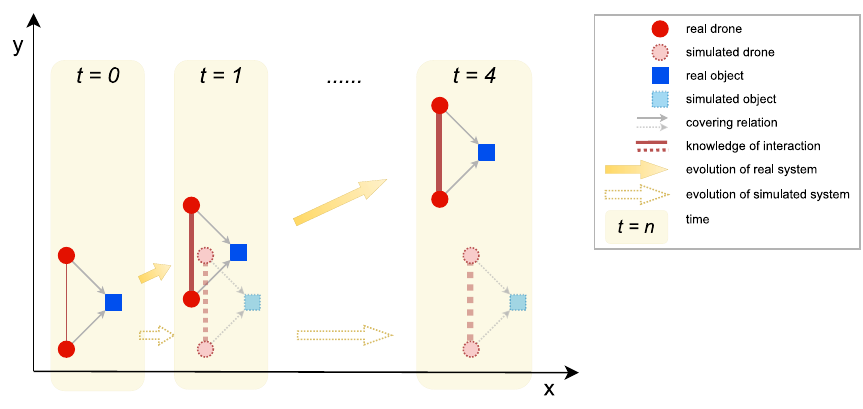}
\caption{A situation where the state deviates but knowledge and utility are not. At t=0, the simulated world and real world are identical, hence they overlap at the same position.}
\label{fig_illust_1}
\end{figure}

Performance of knowledge comparison, as depicted in Fig. \ref{fig_illust_1} reveals a common situation under threats \ref{T_param}: though the utility and knowledge are correctly simulated, the drones of the simulated world show complete deviation, when compared to their real positions. 
In this situation, two drones are tracking one object. In both the real and simulated worlds, the two drones are always covering the given object. The utility of 2-coverage for this two-drone system is always 1 in both worlds during all the 5 time steps. The knowledge in both systems is incremented the same way thus being identical.
However, since the positions of the two systems are different, state comparison will regard the simulation as "deviated" and then will trigger an update, but knowledge comparison will not. Therefore, this situation shows that even with no utility deviation, state comparison can still invoke unnecessary updates, as observed in the highlighted areas of Fig. \ref{fig_sim}.
Nevertheless, the advantage of knowledge comparison is only prominent when the tolerance by threshold is small. With larger thresholds, the simulation will evolve much further over time until the accumulated deviation is greater than the threshold and triggers the next updates. After evolving longer over time, the complex interaction with other drones and objects may lead to an entirely different system topology in the simulation, where the knowledge, state, and utility are all largely deviated. In this case, monitoring state (position) deviation might be more relevant to utility deviation, since the utility of coverage is calculated based on the relative position between objects and agents. This phenomenon is shown by the non-highlighted areas in Fig. \ref{fig_sim}. 
Also, if contrasting conditions I and II, the highlighted areas in II are generally smaller than I. This is because the large inaccuracy of the positions of uncovered objects under condition II causes the knowledge and system topology to deviate faster than condition I. For condition II, only when the threshold is smaller when compared to I, knowledge comparison can outperform the state comparison. 

Therefore, when the requirement of the DT application focuses on minimising the utility deviation of the simulation, knowledge comparison can reduce much more overheads caused by unnecessary updates. With the same number of updates, knowledge comparison also ensures a more accurate simulation of the task utility when compared to state comparison.

We then quantitatively compare the three approaches by the notion of \textit{hypervolume}, which measures the quality of a Pareto front by the size of the space dominated by all nondominated solution points with respect to a reference point \cite{While2006Hypervolume}. A larger hypervolume is better. In this paper, we use $(1, 1)$ as the reference point for hypervolume calculation. 
The results are shown in Table \ref{table_hypervolume}.
The values in the table confirm our observations above for the three approaches: in almost all the cases, knowledge comparison has a larger hypervolume than state comparison; action comparison has the smallest hypervolume.
\begin{table}[!t]
\caption{All solutions measured by the hypervolume of the two objectives: utility deviation and the number of updates.
}
\label{table_hypervolume}
\centering
\scalebox{0.85}{
\begin{tabular}{lcccccc}
\toprule
\multirow{2}{4em}{Methods} & \multicolumn{6}{c}{Test Scenarios} \\\cmidrule{2-7}
 & I-1 & I-2 & I-3 & I-4 & I-5 & I-6\\
\midrule
State  & \underline{0.9503} & \underline{0.9456} & \underline{0.9638} & \underline{0.9401} & \underline{0.9231} & \underline{0.9649}\\
Knowledge & \textbf{0.9629} & \textbf{0.9500} & \textbf{0.9643} & \textbf{0.9489} & \textbf{0.9385} & \textbf{0.9699}\\
Action & 0.9090 & 0.8910 & 0.9336 & 0.9013 & 0.8417 & 0.9104\\
\midrule
 & II-1 & II-2 & II-3 & II-4 & II-5 & II-6\\
\cmidrule{2-7}
State & 0.9525 & 0.9607 & \textbf{0.9721} & 0.9458 & 0.9407 & 0.9562 \\
Knowledge & \textbf{0.9635} & \textbf{0.9638} & 0.9711 & \textbf{0.9542} & \textbf{0.9469} & \textbf{0.9646} \\
\bottomrule
\end{tabular}
}
\end{table}

\begin{figure}[!t]
\subfloat[I-1]{\includegraphics[width=0.3\textwidth]{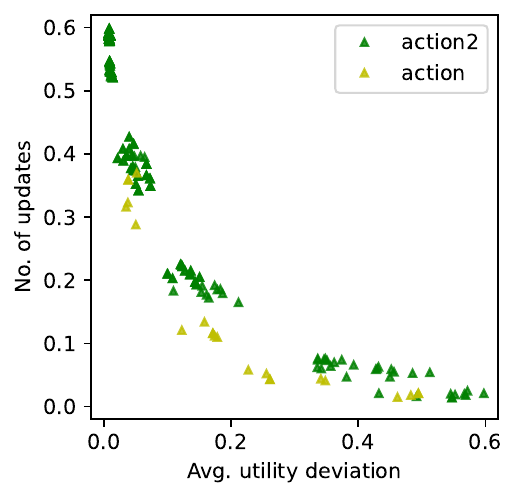}}
\hfil
\subfloat[I-2]{\includegraphics[width=0.3\textwidth]{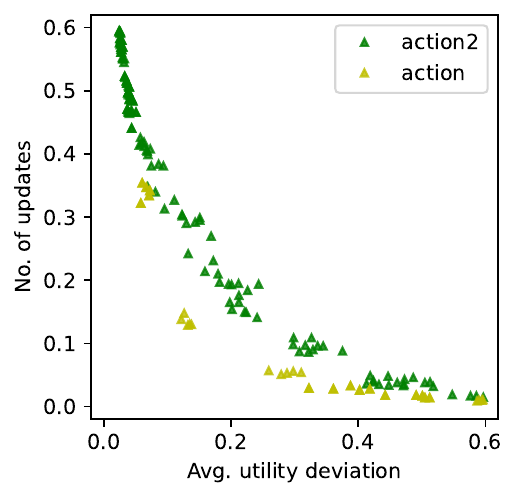}}
\hfil
\subfloat[I-3]{\includegraphics[width=0.3\textwidth]{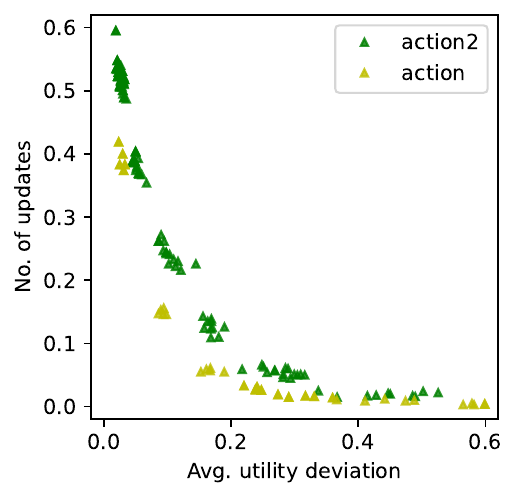}}
\vfil
\subfloat[I-4]{\includegraphics[width=0.3\textwidth]{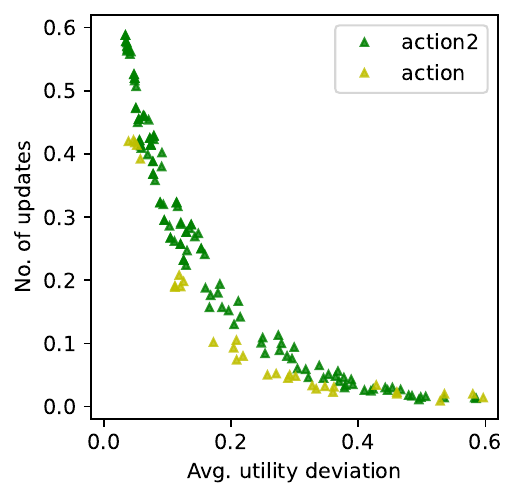}}
\hfil
\subfloat[I-5]{\includegraphics[width=0.3\textwidth]{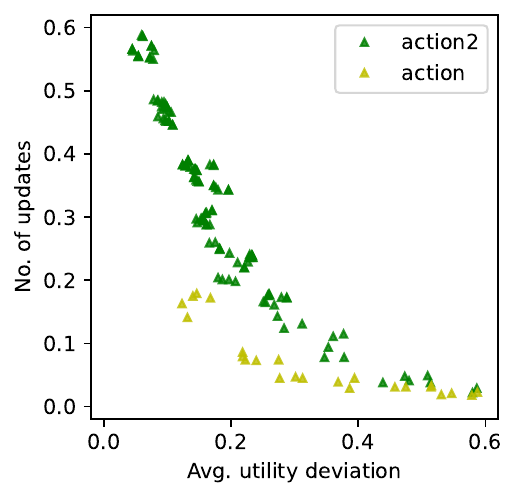}}
\hfil
\subfloat[I-6]{\includegraphics[width=0.3\textwidth]{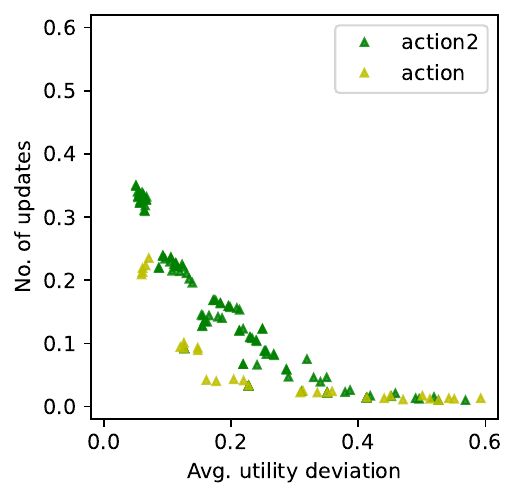}}

\caption{Two different ways of quantifying actions applied to scenarios I-1 to I-6.}
\label{fig: action}
\end{figure}

Additionally, in action comparison, the action and its deviation can be quantified by viewing it at different levels of granularity. The previously evaluated view of the action is regarded as the coarse-grained view.
A fine-grained view of action deviation is then further designed and evaluated. The calculation follows Algorithm \ref{alg: action2}, which defines the deviation of the actions of one physical agent and its virtual agent counterpart at a particular time step. 
The evaluation is based on scenarios I-1 to I-6 and follows the same procedure as Section \ref{sec: pareto evaluation}.
The results are shown in Fig. \ref{fig: action}, in which the newly introduced fine-grained view is denoted as \texttt{action2} and the coarse-grained view is denoted as \texttt{action}.
According to the results, the coarse-grained view \texttt{action} is more Pareto efficient in all 6 scenarios. 
One possible reason is that the \texttt{action} is more related to the utility deviation.
The utility is measured by the number of objects being covered in a global view, not who is covering which specific object.
The four coarse actions used by \texttt{action} indicate whether an agent is covering any important objects, which directly contributes to the utility of k-coverage.
For instance, if the virtual and physical agents are both doing the action \texttt{respond-and-follow}, then this means they have both not sensed any important object. Then \texttt{action} will regard them as equivalent, because responding to which particular drone and trying to follow which object does not make any difference in k-coverage at the current moment.
However, \texttt{action2} would distinguish in detail about the other drone and object, which can be too strict in specifying deviation.

\subsection{Memory Usage of Knowledge Equivalence Checking}

\begin{figure}[!t]
\centering
\includegraphics[width=0.5\linewidth]{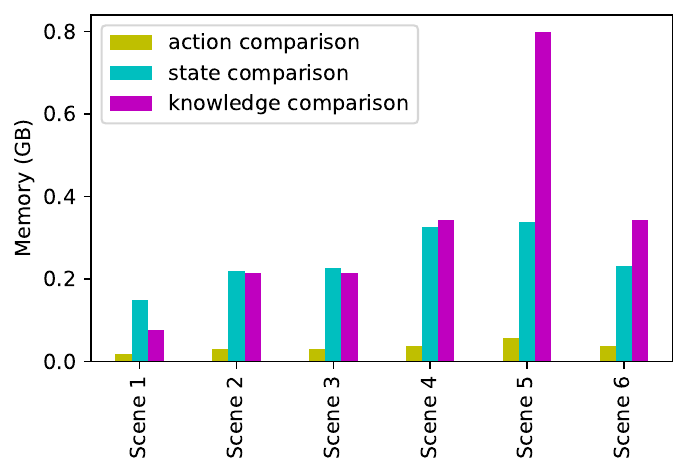}
\caption{Extra memory required for the comparison.}
\label{fig: memory}
\end{figure}

 The memory usage of each method is evaluated and the results are shown in Fig. \ref{fig: memory}. The y-axis of the figure is the total extra amount of sensor data that needs to be loaded into the memory for comparison.
Action comparison uses much less memory when compared to the other two, since an action can be encoded as one integer. Despite being a bit less Pareto efficient than the other two methods as in \figurename \ref{fig_sim}, action comparison is the best choice when less memory consumption is preferred.
Although knowledge comparison is more Pareto efficient in the previous evaluation, its memory consumption will increase exponentially as the graph size increases. For instance, scene 4 has 10 cameras and 45 graph edges, while scene 5 involves 15 cameras and 105 edges, causing memory consumption to double in scene 5 than scene 4.

\subsection{Discussion}

\subsubsection{Evaluation Summary}
The experiments have evaluated the effectiveness of knowledge update, and the performance of the two proposed knowledge equivalence checking approaches in terms of their update overhead, simulation accuracy and memory consumption.
The evaluation is set given different types of threats to equivalence in section \ref{sec:threats}.
The results have answered the two questions at the beginning of this section.
First, for \ref{Q_update}, maintaining knowledge equivalence is essential in keeping the DT model up-to-date, since it on average leads to lower simulation deviation of utility.
Second, for \ref{Q_checking}, the proposed knowledge equivalence checking approaches have demonstrated their performance advances over the baseline (state comparison) in terms of Pareto efficiency.
When a high simulation accuracy on task utility is preferred, knowledge comparison achieves the best trade-offs of reducing update overhead and increasing accuracy. When fewer updates are preferred, the three methods are generally comparable. Although knowledge comparison is sometimes slightly less Pareto efficient than the baseline approach specifically when utility deviation is large (e.g. in I-2 and I-3), such a situation violates the fundamental aim of DT being an accurate representation.
The quantitative results measured by the hypervolume metric also show that knowledge comparison is  generally comparable to or better than the baseline.
Although the proposed action comparison does not perform better than the other two methods in the Pareto efficiency of update overhead vs. accuracy, it can largely reduce memory consumption. Action comparison scales much better than knowledge comparison and state comparison in terms of the number of objects. Action comparison is most suitable in situations where a large number of objects are involved while the number of drones is relatively small.

The proposed knowledge equivalence-checking approaches can also be promising in other application domains.
An indicative example is mobile edge computing. The mobile devices offload some of their computation to the edge servers (e.g. base stations) nearby. Each edge server can leverage self-awareness to accumulate knowledge of the behaviour patterns of the surrounding mobile devices (e.g. latency, urgency of requests, computation capacity of the device) and past interaction with other edge servers. Based on this local knowledge, each edge server can thereby automatically decide whether to accept or migrate the offloaded computation in a load-balanced way.
A DT of the entire edge-device system can provide a global view for more optimised offloading with what-if simulation analysis, but it may be infeasible to keep equivalent environment (mobile devices and their users) states every moment due to highly dynamic user behaviours. 
Nevertheless, knowledge is an edge server's world model and based on which the server takes actions for macro-level objectives (e.g. overall users satisfaction, quality of services). Knowledge has a closer link with these objectives than the state does. Therefore, using knowledge equivalence checking can tolerate state deviations that do not cause deviations in macro-level objectives, thus reducing the frequency of updates needed.

\subsubsection{Threats to Validity}

The experiment's validity can be compromised by several threats that arise from potential bias in the case study and the assumptions made for the experiment design. Identifying and addressing these threats is crucial for ensuring the reliability of the results, and these threats are presented below.

\paragraph{Real-World Data and Case Studies}
We have conducted controlled experiments using simulated data. Simulated data have provided us with a cost-effective path for learning about the behaviour and performance of enacting knowledge equivalence in DT under a range of scenarios that can exemplify complex real settings. Nevertheless, further systematic studies will consider data from real-world systems and domain-specific properties to evaluate for knowledge equivalence. 

The validity of the evaluation case study is made possible with two implicit assumptions: first, minor differences in the state are indistinguishable when abstracted as knowledge, and will not change the behaviour pattern of the drones; and second, knowledge deviation is more sensitive to changes in task utility, compared to state deviation.
Therefore, the state imprecision in the DT can be tolerated to some extent, and will not cause deviation in task utilities between the simulated world and the real world. Differences in knowledge indicate that a prominent deviation possibly has happened in the state, and it is also why the knowledge has drifted.
Only example cases that share the same assumptions above can observe a similar improvement in performance by knowledge equivalence checking.
Even in the evaluation, these two assumptions do not hold all the time, which explains only under the preference for minimising deviation, knowledge comparison can be more Pareto efficient than state comparison.

\paragraph{Types of Threats to Equivalence}
In the real world, different types of uncertainties exist. In our experiments, only three types of threats to equivalence are considered. Further studies are needed to evaluate for the manifestation of  multiple types and/or compounded threats that may exist at the same time, including the threats mentioned in section \ref{sec:threats}. 
In addition, our controlled experiments assume a situation where the configuration of the environment is known: the number of objects and how an object will become important is known. Further studies need to consider highly volatile dynamic and opportunistic scenarios. 

\paragraph{Simplification in the Evaluation Model}
This paper adopts a simplified model for calculating network and computation overhead.
The overhead caused by network communication is not studied. For systems consisting of many drones and objects, data transmission to the DT may become a bottleneck and can increase drone energy consumption.
The time required for data loading and processing in the DT is not modelled. Instead, this paper measures the overhead caused by data loading and processing by the ``number of updates''. To achieve real-time or faster-than-real-time simulation, the model must match the application regarding simulation time scale, model fidelity, and computation efficiency. This paper assumes that the DT can run simulations in real time, but further research is needed to investigate time limitations when running simulations that must meet time constraints for prompt decision-making.

\section{Summary and Future work} \label{sec:conclusion}

This paper has investigated the problem of knowledge equivalence in the context of digital twins (DT) of intelligent multi-agent systems. The objective of the work is to ensure that knowledge equivalence and self-awareness are kept up-to-date in the DT and synchronised with the physical world. Knowledge equivalence is particularly important for unlocking cognitive capabilities in the DT that can promote new opportunities for sophisticated analysis (e.g., what-if prognostic and diagnostic analysis) that can further support and enhance intelligence in the physical world. Additionally, our contribution shows that knowledge can provide an alternative equivalence metric for such a complex setting, where using knowledge equivalence can alleviate overheads of continuous calibration of models, which leverage state, and/or input parameters comparison and updates.  

The paper has presented a conceptual architecture for intelligent self-aware DT and an approach for checking knowledge equivalence at the interaction level. 
Our analysis of the proposed approach demonstrates that    
knowledge equivalence can tolerate deviation thus reducing unnecessary updates and unwanted overheads. When compared to discrepancy analysis based on state, knowledge equivalence can achieve more Pareto efficient results on the trade-off between overhead (number of updates) and effectiveness (utility deviation). This can significantly reduce update overheads, when lower utility deviation is preferred.

Our future work will consider incorporating more sophisticated statistical techniques like hypothesis tests to improve the knowledge discrepancy analysis process. Additionally, we are exploring new self-adaptive mechanisms for selective updates of the physical agent(s) in situations, where intelligence in DT exceeds that of the physical setting and mirroring the updates can be particularly important for harnessing intelligence in the physical world. In the current model, agents are homogeneous in their decision models. Our future work will consider scenarios with  heterogeneous agents with more sophisticated utility functions and metrics that capture different properties of the system under analysis.

\begin{acks}
 This research was supported by: Shenzhen Science and Technology Program,  China (No. GJHZ20210705141807022); SUSTech-University of Birmingham Collaborative PhD Programme; Guangdong Province Innovative and Entrepreneurial Team Programme, China (No. 2017ZT07X386); SUSTech Research Institute for Trustworthy Autonomous Systems, China; and EPSRC/EverythingConnected Network project on Novel Cognitive Digital Twins for Compliance, UK.
\end{acks}

\bibliographystyle{ACM-Reference-Format}
\bibliography{references}

\appendix
\section{Agent Behaviours in the Motivating Example}
\label{Agent behaviour in the motivating Example}
\begin{figure}[!t]
    \removelatexerror
    
    \begin{algorithm}[H]
    \caption{Drone Agent Behaviour.\label{alg: camera}}
    \SetAlgoLined
    \KwData{Knowledge of interaction $graph$, 
    drone sensing range $range$, 
    list of all received messages $msgList$,}

    \For{every time step} {
        \tcc{SENSE}
        $objList \leftarrow$ sense all objects within $range$\;
        $imptObjList \leftarrow$ filter important objects from $objList$\;
        $msgList.append($messages received at previous time step$)$\;
        Clear outdated messages in $msgList$\;
        \tcc{THINK and ACT}
        \uIf{!imptObjList.isEmpty()} {
            $obj \leftarrow$ randomly select from $imptObjList$\;
            \If{!obj.isKCovered()}{
                \tcp*[h]{notify} \\
                $notifiedDrones \leftarrow$ the top $(k-1)$ drones ranked by the strengths of their links with this drone, which are recorded as the edge weights in $graph$\;
                \ForEach{drone $\in$ notifiedDrones}{
                    Send message to $drone$\;
                }
            }
            \tcp*[h]{follow} \\
            Move towards $obj$ by 1 unit\;
        }
        
        \uElseIf{!msgList.isEmpty()} {
            \tcp*[h]{respond-and-follow} \\
            Sort messages in $msgList$ in descending order first by the weight of edge between the sender drone and this drone, then by the receiving time\;
            $msg \leftarrow$ the message at the highest rank after sorting\;
            $obj \leftarrow$ the object advertised in $msg$\;
            Move towards $obj$ by 2 unit\;
        }
        \Else(\tcp*[h]{random walk}){
            Move towards a random angle by 5 units\;
        }
        \tcc{UPDATE KNOWLEDGE}
        \ForEach{edge $\in$ graph.allEdges}{
            $edge.weight \leftarrow \gamma \cdot edge.weight$\;
        }
        \ForEach{obj $\in$ $ObjList$}{
            \ForEach{d $\in$ all surrounding drones}{
                \If{d.isCovering(obj)}{
                    $edge \leftarrow$ the graph edge between drone $d$ and this drone\;
                    $edge.weight \leftarrow edge.weight + \delta$\;
                }
            }
        }
     }
    \end{algorithm}

\end{figure}

\end{document}